\documentclass[journal,twoside,web]{ieeecolor}
\usepackage{tmi}
\usepackage{cite}
\usepackage{amsmath,amssymb,amsfonts}
\usepackage{algorithmic}
\usepackage[dvipsnames]{xcolor}
\usepackage{graphicx}
\usepackage{textcomp}
\usepackage{float} 
\usepackage{booktabs}
\usepackage{bm}
\usepackage{algorithm}
\usepackage{pgfplots}
\usepackage{mathtools}
\usepackage{url}
\usepackage{cite}
\usepackage[section]{placeins}
\usepackage{subfig}
\usepackage{makecell}
\usepackage{multirow}
\usepackage{multicol}
\usepackage{epstopdf}

\newcommand{\etal}{\textit{et al.}}

\makeatletter
\newcommand{\printfnsymbol}[1]{\textsuperscript{\@fnsymbol{#1}}}
\makeatother
\newcommand{\PreserveBackslash}[1]{\let\temp=\\#1\let\\=\temp}
\def\BibTeX{{\rm B\kern-.05em{\sc i\kern-.025em b}\kern-.08em
    T\kern-.1667em\lower.7ex\hbox{E}\kern-.125emX}}
\begin{document}
\title{nnFormer: Volumetric Medical Image Segmentation via a 3D Transformer}
\author{Hong-Yu Zhou, \IEEEmembership{Student Member, IEEE}, Jiansen Guo, Yinghao Zhang, Xiaoguang Han, \\ Lequan Yu, Liansheng Wang, \IEEEmembership{Member, IEEE}, and Yizhou Yu, \IEEEmembership{Fellow, IEEE}
\thanks{(\emph{Corresponding author: Liansheng Wang and Yizhou Yu.})}
\thanks{This work was done when Hong-Yu Zhou was a visiting student at Xiamen University.}
\thanks{Hong-Yu Zhou, Jiansen Guo, Yinghao Zhang and Liansheng Wang are with the Department of Computer Science, Xiamen University, Siming District, Xiamen, Fujian Province, P.R. China (email: whuzhouhongyu@gmail.com, jsguo@stu.xmu.edu.cn, zhangyinghao@stu.xmu.edu.cn, lswang@xmu.edu.cn).}
\thanks{Hong-Yu Zhou and Yizhou Yu are with the Department of Computer Science, The University of Hong Kong, Pokfulam, Hong Kong (e-mail: yizhouy@acm.org).}
\thanks{Xiaoguang Han is with the Shenzhen Research Institute of Big Data, The Chinese University of Hong Kong (Shenzhen), Shenzhen, Guangdong Province, P.R. China (email: hanxiaoguang@cuhk.edu.cn).}
\thanks{Lequan Yu is with the Department of Statistics and Actuarial Science, The University of Hong Kong, Pokfulam, Hong Kong (e-mail: lqyu@hku.hk).}
\thanks{\emph{First two authors contributed equally.}}
}
\maketitle
\begin{abstract}
Transformer, the model of choice for natural language processing, has drawn scant attention from the medical imaging community. Given the ability to exploit long-term dependencies, transformers are promising to help atypical convolutional neural networks to overcome their inherent shortcomings of spatial inductive bias. However, most of recently proposed transformer-based segmentation approaches simply treated transformers as assisted modules to help encode global context into convolutional representations. To address this issue, we introduce nnFormer (i.e., \textbf{n}ot-a\textbf{n}other trans\textbf{Former}), a 3D transformer for volumetric medical image segmentation. nnFormer not only exploits the combination of interleaved convolution and self-attention operations, but also introduces local and global volume-based self-attention mechanism to learn volume representations. Moreover, nnFormer proposes to use skip attention to replace the traditional concatenation/summation operations in skip connections in U-Net like architecture. Experiments show that nnFormer significantly outperforms previous transformer-based counterparts by large margins on three public datasets. Compared to nnUNet, nnFormer produces significantly lower HD95 and comparable DSC results. Furthermore, we show that nnFormer and nnUNet are highly complementary to each other in model ensembling. Codes and models of nnFormer are available at \url{https://git.io/JSf3i}. 
\end{abstract}


\section{Introduction}
\label{sec:introduction}
Transformer~\cite{vaswani2017attention}, which has become the de-facto choice for natural language processing (NLP) problems, has recently been widely exploited in vision-based applications~\cite{dosovitskiy2020image,liu2021swin,he2021masked,carion2020end}. The core idea behind is to apply the self-attention mechanism to capture long-range dependencies. Compared to convolutional neural networks (i.e., convnets~\cite{lecun1998gradient}), transformer relaxes the inductive bias of locality, making it more capable of dealing with non-local interactions \cite{zhou2021convnets,qu2022m3net,zhang2021cross}. It has also been investigated that the prediction errors of transformers are more consistent with those of humans than convnets \cite{tuli2021convolutional}.

Given the fact that transformers are naturally more advantageous than convnets, there are a number of approaches trying to apply transformers to the field of medical image analysis. Chen \etal~\cite{chen2021transunet} first time proposed TransUNet to explore the potential of transformers in the context of medical image segmentation. The overall architecture of TransUNet is similar to that of U-Net \cite{ronneberger2015u}, where convnets act as feature extractors and transformers help encode the global context. In fact, one major characteristic of TransUNet and most of its followers \cite{zhang2021transfuse,valanarasu2021medical,chang2021transclaw,chen2021transattunet} is to treat convnets as main bodies, on top of which transformers are further applied to capture long-term dependencies. However, such feature may cause a problem, which is the advantages of transformers are not fully exploited. In other words, we believe one- or two-layer transformers are not enough to entangle long-term dependencies with convolutional representations that often contain precise spatial information and provide hierarchical concepts.

To address the above issue, some researchers \cite{karimi2021convolution,cao2021swin,lin2021ds} started to use transformers as the main stem in segmentation models. Karimi \etal~\cite{karimi2021convolution} first time introduced a convolution-free segmentation model by forwarding flattened image representations to transformers, whose outputs are then reorganized into 3D tensors to align with segmentation masks. Recently, Swin Transformer~\cite{liu2021swin} showed that by referring to the feature pyramids used in convnets, transformers can learn hierarchical object concepts at different scales by applying appropriate down-sampling to feature maps. Inspired by this idea, SwinUNet \cite{cao2021swin} utilized hierarchical transformer blocks to construct the encoder and decoder within a U-Net like architecture, based on which DS-TransUNet~\cite{lin2021ds} added one more encoder to accept different-sized inputs. Both SwinUNet and DS-TransUNet have achieved consistent improvements over TransUNet. Nonetheless, they did not explore how to appropriately combine convolution and self-attention for building an optimal medical segmentation network. 

\begin{figure}[t]
    \centering
    \includegraphics[width=1.0\columnwidth]{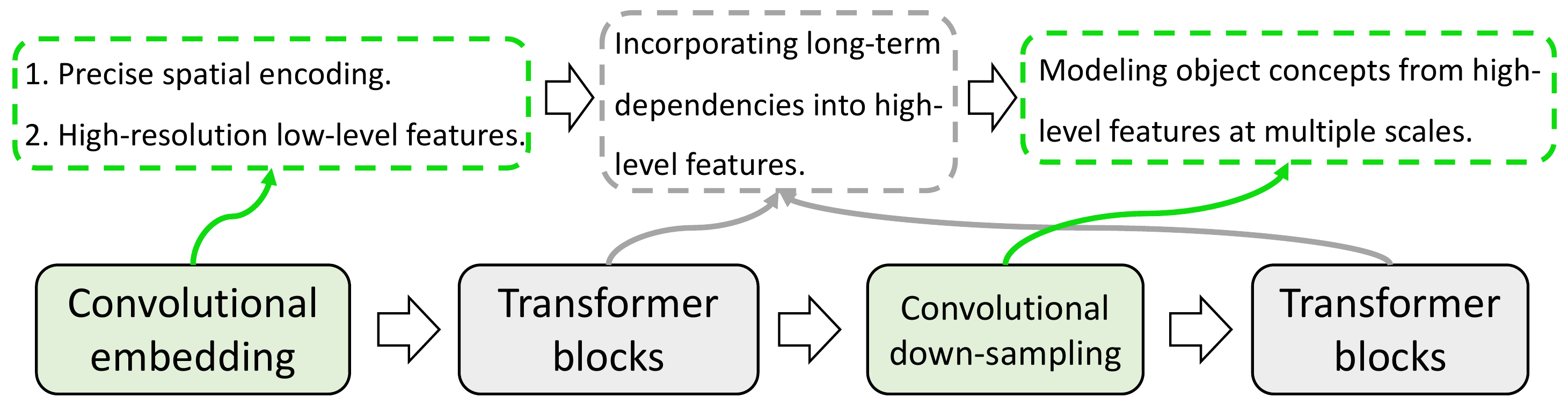}
    \caption{The interleaved stem used in the encoder of nnFormer.}
    \label{stem}
\end{figure}

In contrast, nnFormer (i.e., \textbf{n}ot-a\textbf{n}other trans\textbf{Former}) uses a hybrid stem where convolution and self-attention are interleaved to give full play to their strengths. Figure \ref{stem} presents the effects of different components used in the encoder of nnFormer. Firstly, we put a light-weight convolutional embedding layer ahead of transformer blocks. In comparison to directly flattening raw pixels and applying 1D pre-processing in \cite{karimi2021convolution}, the convolutional embedding layer encodes precise (i.e., pixel-level) spatial information and provides low-level yet high-resolution 3D features. After the embedding block, transformer and convolutional down-sampling blocks are interleaved to fully entangle long-term dependencies with high-level and hierarchical object concepts at various scales, which helps improve the generalization ability and robustness of learned representations. 

The other contribution of nnFormer lies in proposing a computational-efficient way to leverage inter-slice dependencies. To be specific, nnFormer proposes to jointly use Local Volume-based Multi-head Self-attention (LV-MSA) and Global Volume-based Multi-head Self-attention (GV-MSA) to construct feature pyramids and provide sufficient receptive field for learning representations on both local and global 3D volumes, which are then aggregated to make predictions. Compared to the naive multi-head self-attention (MSA) \cite{vaswani2017attention}, the proposed strategy can greatly reduce the computational complexity while producing competitive segmentation performance. Moreover, inspired by the attention mechanism used in the task of machine translation~\cite{vaswani2017attention}, we introduce skip attention to replace the atypical concatenation/summation operation in skip connections of U-Net like architecture, which further improves the segmentation results.

% In the experiment section, we first compare nnFormer with a wide range of transformer-based models, where nnFormer surpasses the previous best performing baseline by substantial and significant margins in both HD95 and DSC metrics on three well-established public datasets. Next, we investigate the significance of nnFormer over nnUNet. The experimental results demonstrate that compared to nnUNet, our nnFormer is more advantageous under HD95 while showing comparable performance under DSC. More importantly, we found that simply averaging the predictions of nnFormer and nnUNet can further bring large improvements on all three public datasets, which suggests that either nnFormer or nnUNet can be treated as an optimal complement to each other in model ensembling. 

To sum up, our contributions can be summarized as follows:
\begin{itemize}
    \item We introduce nnFormer, a 3D transformer for volumetric medical image segmentation. nnFormer achieves significant improvements over previous transformer-based medical segmentation models on three well-established datasets.
    \item Technically, the contributions of nnFormer are three folds: i) an interleaved combination of convolution and self-attention operations. ii) the utilization of both local and global volume-based self-attention to build feature pyramids and provide large receptive fields, respectively. iii) skip attention is proposed to replace traditional concatenation/summation operations in skip connections.
    \item Thorough experiments have been conducted to validate the advantages of nnFormer over nnUNet. We show that nnFormer is significantly better than nnUNet in hausdorff distance and achieves slightly better performance in dice coefficient. Moreover, we found that nnFormer and nnUNet are highly complementary to each other as simply averaging their predictions can already greatly boost the overall performance.
\end{itemize}

%The proposed nnFormer surpasses Swin-UNet by over 7 percents in the task of multi-organ segmentation on Synapse. When performing automated cardiac diagnosis on ACDC dataset, nnFormer outperforms Swin-UNet by nearly 2 percents in average. Considering the average dice score on ACDC is over 90 percents, we believe 2-percent improvements on ACDC are as impressive as the 7-percent improvements on Synapse. 

\section{Related Work}
\label{sec:relatedwork}
In this section, we mainly review methodologies that resort to transformers to improve segmentation results of medical images. Since most of them employ hybrid architecture of convolution and self-attention \cite{vaswani2017attention}, we divide them into two categories based on whether the majority of the stem is convolutional or transformer-based.\\

\noindent \textbf{Convolution-based stem.} TransUNet \cite{chen2021transunet} first time applied transformer to improve the segmentation results of medical images. TransUNet treats the convnet as a feature extractor to generate a feature map for the input slice. Patch embedding is then applied to patches of feature maps in the bottleneck instead of raw images in ViT \cite{dosovitskiy2020image}. Concurrently, similar to TransUNet, Li \etal \cite{li2021medical} proposed to use a squeezed attention block to regularize the self-attention modules of transformers and an expansion block to learn diversified representations for fundus images, which are all implemented in the bottleneck within convnets. TransFuse \cite{zhang2021transfuse} introduced a BiFusion module to fuse features from the shallow convnet-based encoder and transformer-based segmentation network to make final predictions on 2D images. Compared to TransUNet, TransFuse mainly applied the self-attention mechanism to the input embedding layer to improve segmentation models on 2D images. Yun \etal \cite{yun2021spectr} employed transformers to incorporate spectral information, which are entangled with spectral information encoded by convolutional features to address the problem of hyperspectral pathology. Xu \etal \cite{xu2021levit} extensively studied the trade-off between transformers and convnets and proposed a more efficient encoder named LeViT-UNet. Li \etal \cite{li2021more} presented a new up-sampling approach and incorporated it into the decoder of UNet to model long-term dependencies and global information for better reconstruction results. TransClaw U-Net \cite{chang2021transclaw} utilized transformers in UNet with more convolutional feature pyramids. TransAttUNet \cite{chen2021transattunet} explored the feasibility of applying transformer self attention with convolutional global spatial attention. Xie \etal \cite{xie2021cotr} adopted transformers to capture long-term dependencies of multi-scale convolutional features from different layers of convnets. TransBTS \cite{wang2021transbts} first utilized 3D convnets to extract volumetric spatial features and down-sample the input 3D images to produce hierarchical representations. The outputs of the encoder in TransBTS are then reshaped into a vector (i.e. token) and fed into transformers for global feature modeling, after which an ordinary convolutional decoder is appended to up-sample feature maps for the goal of reconstruction. Different from these approaches that directly employ convnets as feature extractors, our nnFormer functionally relies on convolutional and transformer-based blocks, which are interleaved to take advantages of each other.\\

\noindent \textbf{Transformer-based stem.} Valanarasu \etal \cite{valanarasu2021medical} proposed a gated axial-attention model (i.e., MedT) which extends the existing convnet architecture by introducing an summational control mechanism in the self-attention. Karimi \etal \cite{karimi2021convolution} removed the convolutional operations and built a 3D segmentation model based on transformers. The main idea is to first split the local volume block into 3D patches, which are then flattened and embedded to 1D sequences and passed to a ViT-like backbone to extract representations. SwinUNet \cite{cao2021swin} built a U-shape transformer-based segmentation model on top of transformer blocks in \cite{liu2021swin}, where observable improvements were achieved. DS-TransUNet \cite{lin2021ds} further extended SwinUNet by adding one more encoder to handle multi-scale inputs and introduced a fusion module to effectively establish global dependencies between features of different scales through the self-attention mechanism. Compared to these transformer-based stems, nnFormer inherits the superiority of convolution in encoding precise spatial information and producing hierarchical representations that help model object concepts at various scales.

\section{Method}
\label{sec:method}
\subsection{Overview}
\begin{figure*}[htp]
  \centering
  \includegraphics[width=2.0\columnwidth]{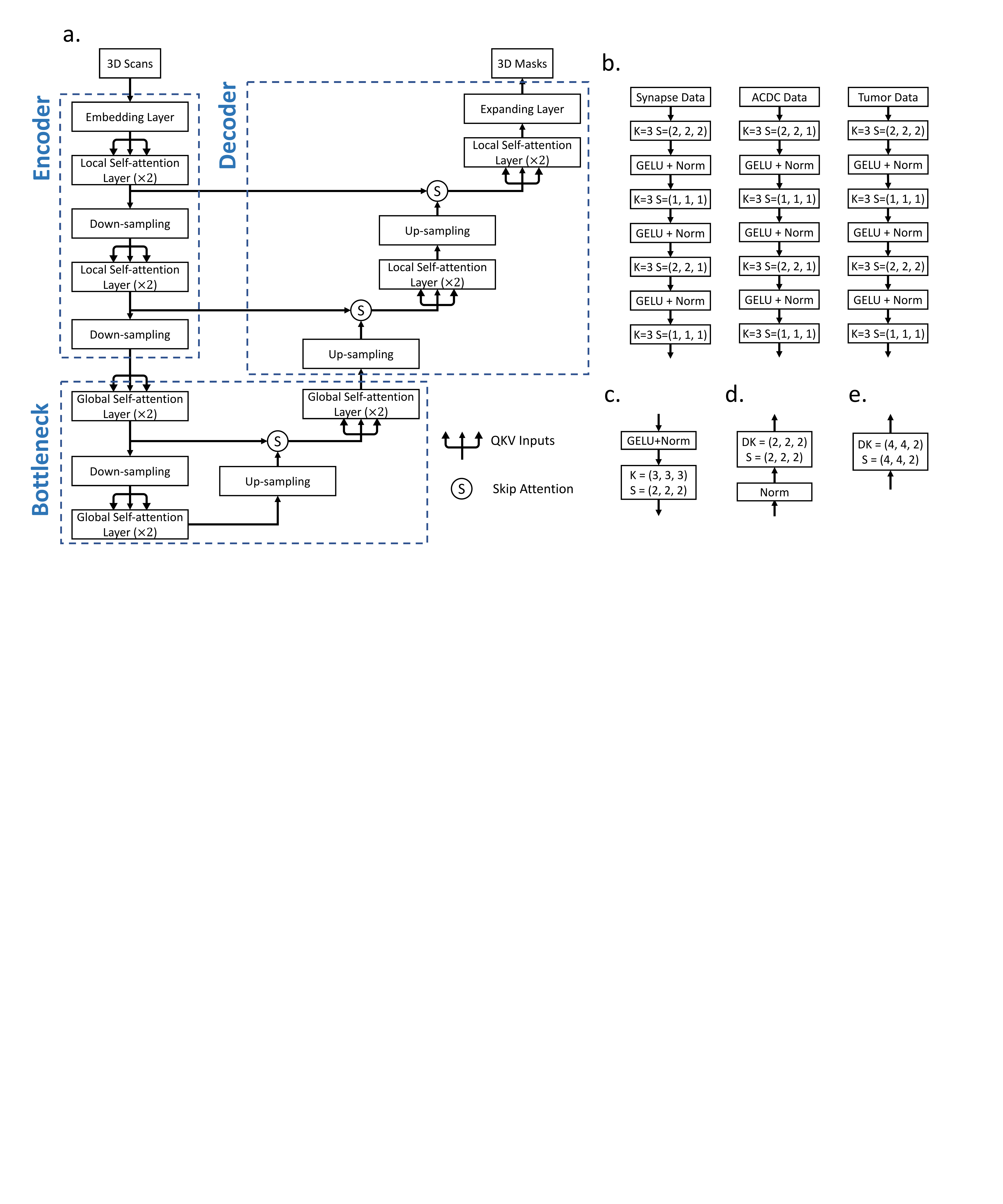}
  \caption{Architecture of nnFormer. In (a), we show the overall architecture of nnFormer. In (b), we present more details of the embedding layers on three publicly available datasets. In (c), (d), (e), we display how to implement the down-sampling, up-sampling and expanding layers, respectively. In practice, the architecture may slightly vary depending on the input scan size. In (b)-(e), \textbf{K} denotes the convolutional kernel size, \textbf{DK} stands for the deconvolutional kernel size and \textbf{S} represents the stride. \textbf{Norm} refers to the layer normalization strategy.} 
  \label{overview}
\end{figure*}

The overall architecture of nnFormer is presented in Figure \ref{overview}, which maintains a similar U shape as that of U-Net \cite{ronneberger2015u} and mainly consists of three parts, i.e., the encoder, bottleneck and decoder. Concretely, the encoder involves one embedding layer, two local transformer blocks (each block contains two successive layers) and two down-sampling layers. Symmetrically, the decoder branch includes two transformer blocks, two up-sampling layers and the last patch expanding layer for making mask predictions. Besides, the bottleneck comprises one down-sampling layer, one up-sampling layer and three global transformer blocks for providing large receptive field to support the decoder. Inspired by U-Net \cite{ronneberger2015u}, we add skip connections between corresponding feature pyramids of the encoder and decoder in a symmetrical manner, which helps to recover fine-grained details in the prediction. However, different from atypical skip connections that often use summation or concatenation operation, we introduce skip attention to bridge the gap between the encoder and decoder. 

In the following, we will demonstrate the forward procedure on Synapse. The forward pass on different datasets can be easily inferred based on the procedure on Synapse.

% The overall architecture of the proposed nnFormer is presented in Figure \ref{fig:1}, which consists the modules including encoder, bottleneck and decoder. Concreatly, the encoder is composed of embedding block x1, swin transformer block x3 and patch merging block x3. The bottle neck is two Swin transformer blocks using W-MSA only. And the decoder is symmetric with the encoder almostly.To achieve the goals of restoring the resolution that reduced in the encoder step by step, The patch merging block and embedding block in the decoder is replaced by patch expanding block and a sole ConvTranspose3d. Each token after the patch merging of the encoder will be added to the decoder. The result of the summation will be fed into the Swin transformer block at the decoder after going through the patch expanding block. 

\subsection{Encoder}
The input of nnFormer is a 3D patch $\mathcal{X} \in \mathcal{R}^{H \times W \times D}$ (usually randomly cropped from the original image), where $H$, $W$ and $D$ denote the height, width and depth of each input scan, respectively.\\

\begin{figure*}[t]
    \centering
    \resizebox{1.0\textwidth}{!}{
    \subfloat[LV-MSA]{
    \includegraphics[width=0.64\columnwidth]{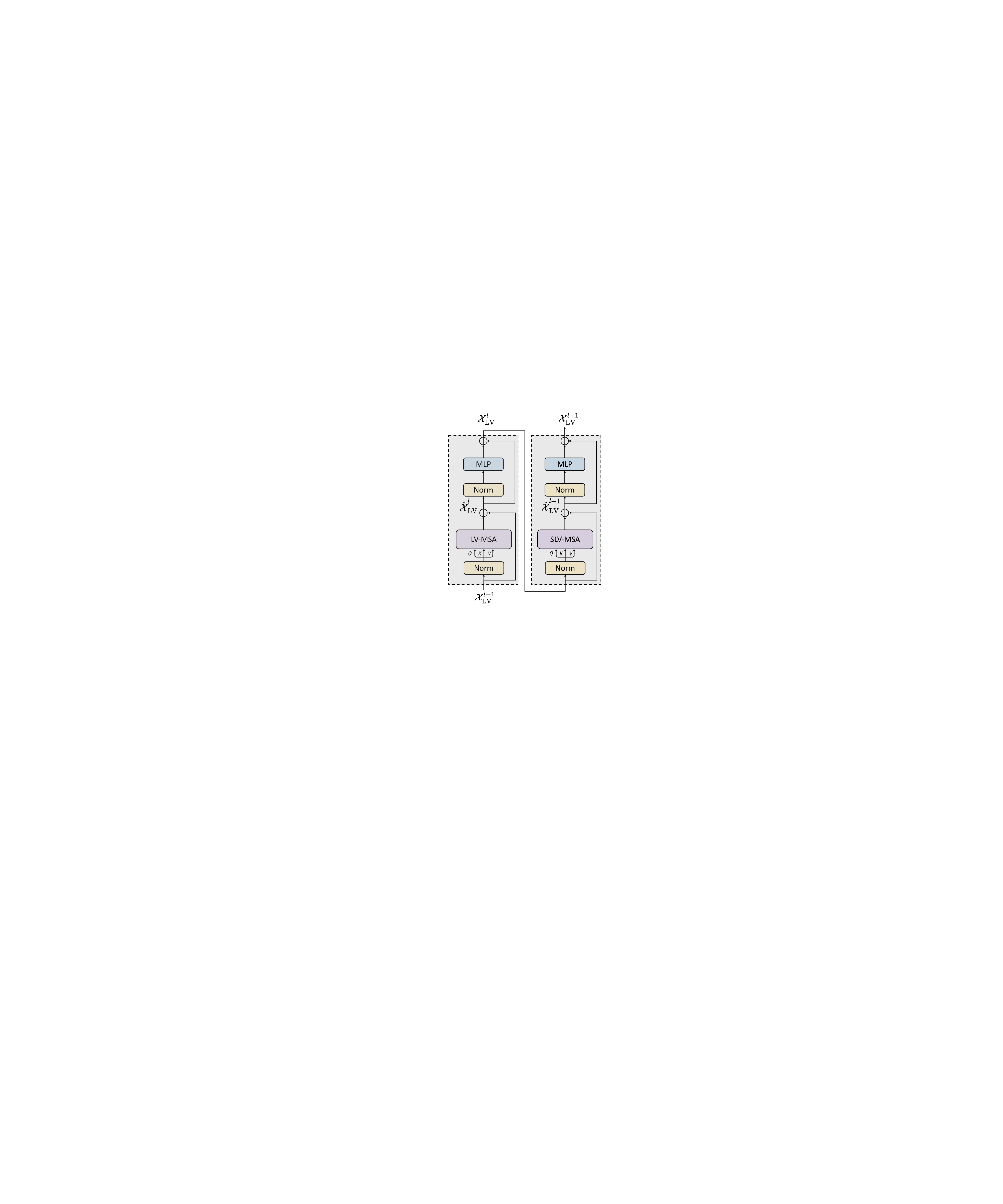}
    \label{fig_lv-msa}
    }\quad
    \subfloat[GV-MSA]{
    \includegraphics[width=0.627\columnwidth]{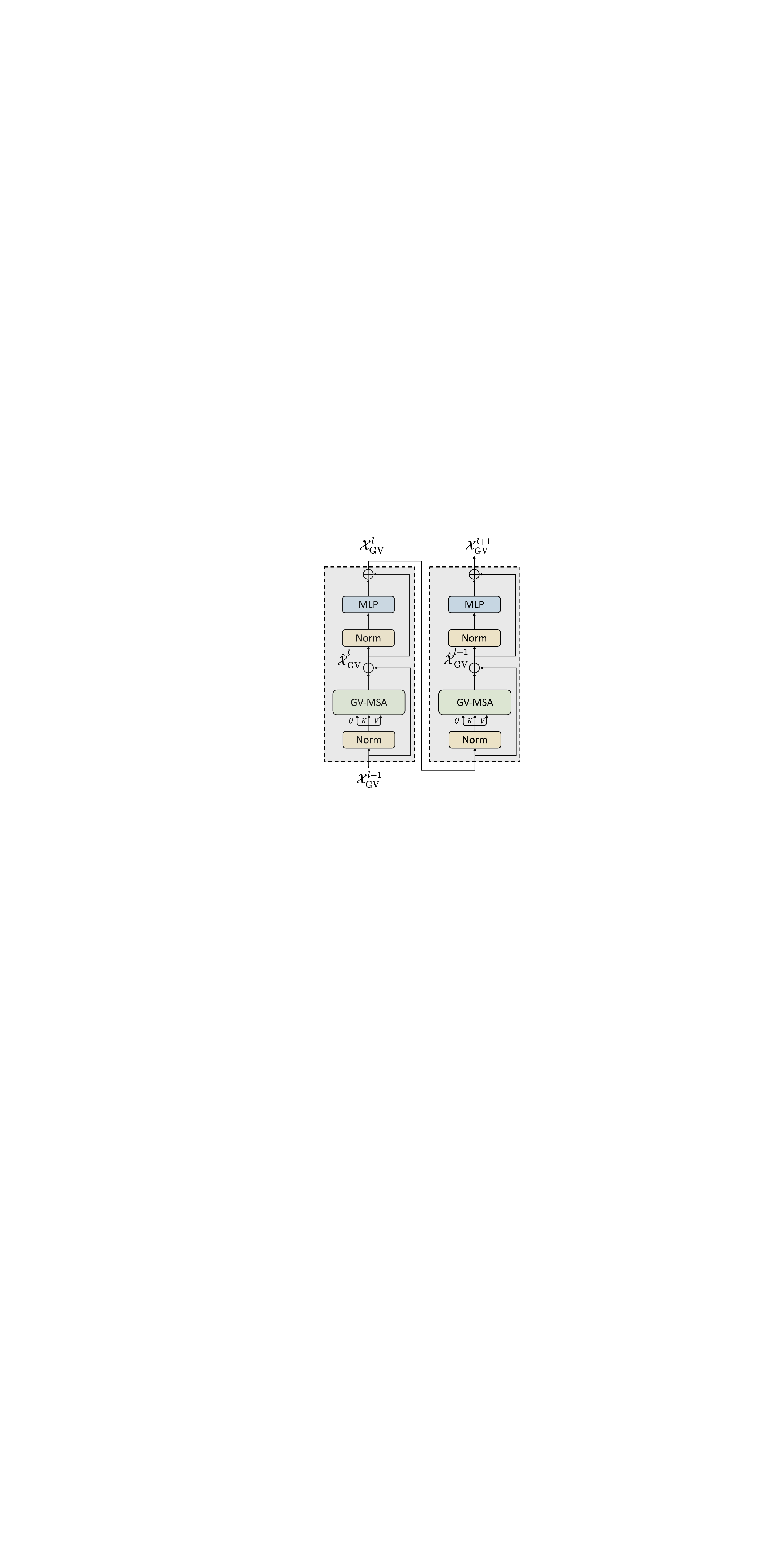}
    \label{fig_gv-msa}
    }\quad
    \subfloat[Skip Attention]{
    \includegraphics[width=0.6\columnwidth]{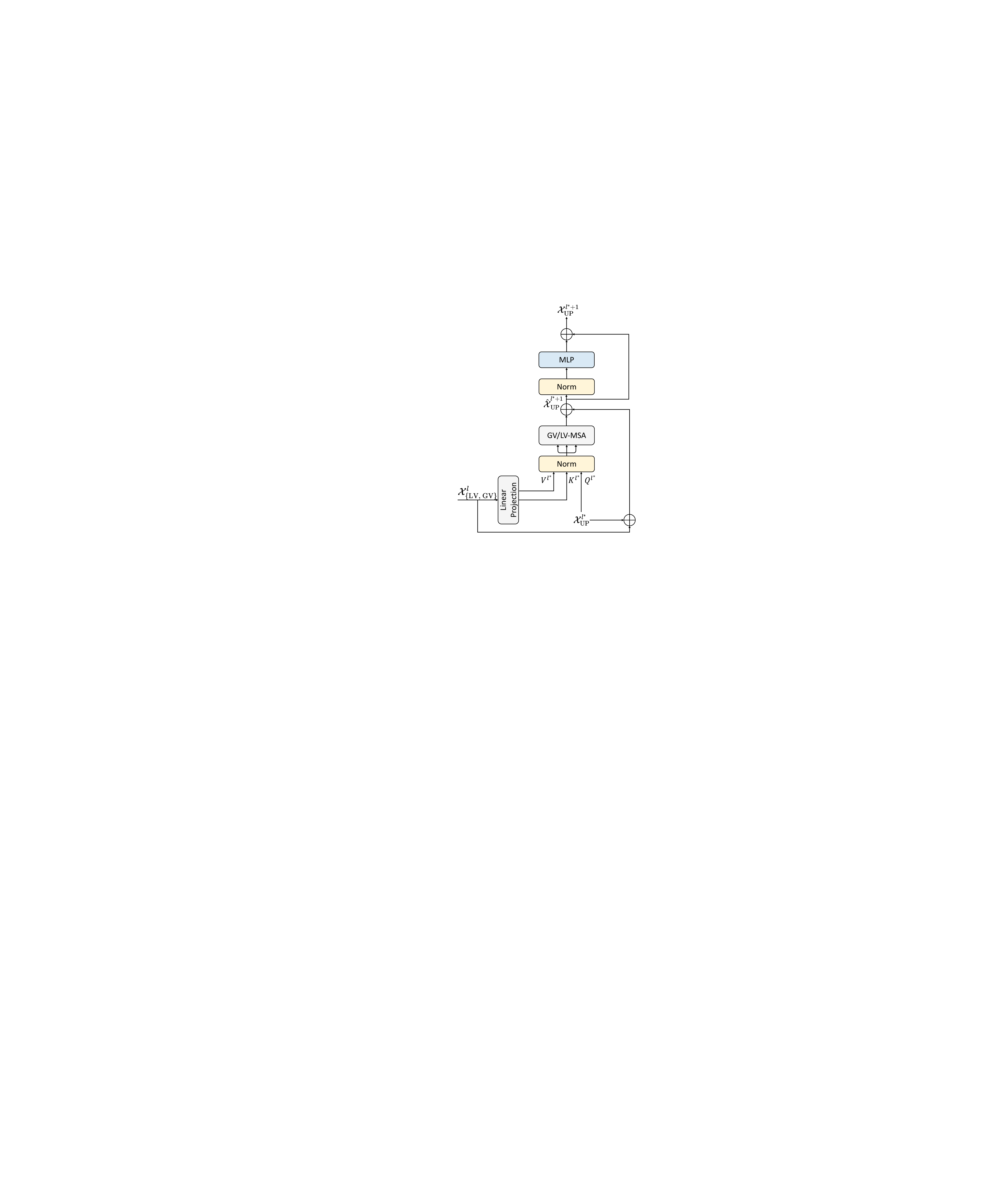}
    \label{fig_skipatt}
    }
    }
    \caption{Three types of attention mechanism in nnFormer. \textbf{Norm} denotes the layer normalization method. \textbf{MLP} is the abbreviation for multi-layer perceptron, which is a two-layer neural network in practice.}
    \label{att_mech}
\end{figure*}

\noindent \textbf{The embedding layer.} On Synapse, the embedding block is responsible for transforming each input scan $\mathcal{X}$ into a high-dimensional tensor $\mathcal{X}_e \in \mathrm{R}^{\frac{H}{4} \times \frac{W}{4} \times \frac{D}{2} \times C}$, where $\frac{H}{4} \times \frac{W}{4} \times \frac{D}{2}$ represents the number of the patch tokens and C represents the sequence length (these numbers may slightly vary on different datasets). Different from ViT \cite{dosovitskiy2020image} and Swin Transformer \cite{liu2021swin} that use large convolutional kernels in the embedding block to extract features, we found that applying {successive} convolutional layers with {small} convolutional kernels bring more benefits in the initial stage, which could be explained from two perspectives, i.e., i) {why applying successive convolutional layers} and ii) {why using small-sized kernels}. For i), we use convolutional layers in the embedding block because they encode pixel-level spatial information, more precisely than patch-wise positional encoding used in transformers. For ii), compared to large-sized kernels, small kernel sizes help reduce computational complexity while providing equal-sized receptive field. As shown in Figure \ref{overview}b, the embedding block consists of four convolutional layers whose kernel size is 3. After each convolutional layer (except the last one), one GELU \cite{hendrycks2016gaussian} and one layer normalization \cite{ba2016layer} layers are appended. In practice, depending on the size of input patch, strides of convolution in the embedding block may accordingly vary.\\

\noindent \textbf{Local Volume-based Multi-head Self-attention (LV-MSA).} After the embedding layer, we pass the high-dimensional tensor $\mathcal{X}_e$ to transformer blocks. The main point behind is to fully {entangle the captured long-term dependencies} with {the hierarchical object concepts at various scales produced by the down-sampling layers} and {the high-resolution spatial information encoded by the initial embedding layer}. Compared to Swin Transformer \cite{liu2021swin}, we compute self-attention within 3D local volumes (i.e., LV-MSA, Local Volume-based Multi-head Self-attention) instead of 2D local windows. 

Suppose that $\mathcal{X}_{\text{LV}} \in \mathcal{R}^{L \times C}$ represents the input of the local transformer block, $\mathcal{X}_{\text{LV}}$ would be first reshaped to $ \mathcal{\hat{X}}_{\text{LV}}\in \mathbf{R}^{N_{\text{LV}} \times N_T \times C}$, where $N_{\text{LV}}$ is a pre-defined number of 3D local volumes and $N_T=S_H\times S_W\times S_D$ denotes the number of patch tokens in each volume. $\{S_H,\ S_W,\ S_D\}$ stand for the size of local volume. 

%In nnFormer, to adapt to various shape of 3D scans, we design $\{S_H,\ S_W,\ S_D\}$ to make it cover all patch tokens of the output of the last transformer block in the encoder. The intuition behind is that it may not be desirable to brute-forcely pad the data in order to satisfy fixed $\{S_H,\ S_W,\ S_D\}$. Thus, the size of the input cropped patch $\mathcal{X}$ needs to adaptively adjusted in order to accord with the size of local volumes. In practice, we set $\{S_H,\ S_W,\ S_D\}$ on Synapse and ACDC to $\{4,4,4\}$ and $\{5,5,3\}$, respectively.

As shown in Figure~\ref{fig_lv-msa}, we follow \cite{liu2021swin} to conduct two successive transformer layers in each block, where the second layer can be regarded as a shifted version of the first layer (i.e., SLV-MSA). The main difference lies in that our computation is built on top of 3D local volumes instead of 2D local windows. The computational procedure can be summarized as follows:
\begin{align}
\small
\begin{split}
    \mathcal{\hat{X}}_{\text{LV}}^{l}&=\text{LV-MSA}\left(\text{Norm}\left(\mathcal{X}_{\text{LV}}^{l-1}\right)\right)+\mathcal{X}_{\text{LV}}^{l-1},\\
    \mathcal{X}_{\text{LV}}^{l}&=\text{MLP}\left(\text{Norm}\left(\mathcal{\hat{X}}_{\text{LV}}^{l}\right)\right)+\mathcal{\hat{X}}_{\text{LV}}^{l},\\
    \mathcal{\hat{X}}_{\text{LV}}^{l+1}&=\text{SLV-MSA}\left(\text{Norm}\left(\mathcal{X}_{\text{LV}}^{l}\right)\right)+\mathcal{X}_{\text{LV}}^{l},\\
    \mathcal{X}_{\text{LV}}^{l+1}&=\text{MLP}\left(\text{Norm}\left(\mathcal{\hat{X}}_{\text{LV}}^{l+1}\right)\right)+\mathcal{\hat{X}}_{\text{LV}}^{l+1}.\\
\end{split}
\end{align}
Here, $l$ stands for the layer index. MLP is an abbreviation for multi-layer perceptron. The computational complexity of LV-MSA on a volume of $h\times w\times d$ patches is:
\begin{align}
    \begin{split}
        \Omega(\text{LV-MSA}) =  4hwd C^2 + 2S_H S_W S_D h w d C.
    \end{split}
    \label{formula_lv-msa}
\end{align}
% Compared to the complexity of the naive multi-head self-attention (MSA) \cite{vaswani2017attention} used in ViT \cite{dosovitskiy2020image}, i.e.,
% \begin{align}
%     \begin{split}
%         \Omega(\text{MSA}) = 4hwd C^2 + 2(hwd)^2 C.
%     \end{split}
% \end{align}
% LV-MSA greatly reduces the computational complexity when dealing with large inputs. For instance, on all four datasets, LV-MSA only needs about 10\% computational resources.

SLV-MSA displaces the 3D local volume used in LV-MSA by $\left(\left\lfloor\frac{S_H}{2}\right\rfloor,\ \left\lfloor\frac{S_W}{2}\right\rfloor,\ \left\lfloor\frac{S_D}{2}\right\rfloor \right)$ to introduce more interactions between different local volumes. In practice, SLV-MSA has the similar computational complexity as that of LV-MSA.

The query-key-value (QKV) attention \cite{vaswani2017attention} in each 3D local volume can be computed as follows:
\begin{align}
    \begin{split}
        \text{Attention}(Q, K, V)=\text{softmax} \left(\frac{Q K^{T}}{\sqrt{d_k}}+B\right) V,
    \end{split}
    \label{qkv_op}
\end{align}
where $Q,K,V \in \mathcal{R}^{N_T\times d_k}$ denote the query, key and value matrices. $B \in \mathcal{R}^{N_T}$ is the relative position encoding. In practice, we first initialize a smaller-sized position matrix $\hat{B} \in \mathcal{R}^{\left(2 S_{H}-1\right) \times\left(2 S_{W}-1\right) \times\left(2 S_{D}-1\right)}$ and take corresponding values from $\hat{B}$ to build a larger position matrix $B$.\\

% It is worth noting that the last transformer block of the encoder (the downmost block in Figure \ref{overview}) only employs V-MSA as we found introducing SV-MSA would deteriorate the overall segmentation results.\\

\begin{table}[t]
     \centering
     \subfloat[][Tumor]
     {
        \scriptsize
    	\begin{tabular}{ccc} \toprule
        &nnFormer& nnUNet\\
        \hline
        Spacing &$[1.0,\ 1.0,\ 1.0]$&$[1.0,\ 1.0,\ 1.0]$\\
        Median shape &$138\times170\times138$&$138\times170\times138$\\
        Crop size &$128\times128\times128$&$128\times128\times128$\\
        Batch size &2&2\\
        DS Str. &\makecell[c]{$[2,2,2],[2,2,2],[2,2,2]$,\\$[2,2,2],[2,2,2]$} &\makecell[c]{$[2,2,2],[2,2,2],[2,2,2]$,\\$[2,2,2],[2,2,2]$}\\
        \bottomrule
    	\end{tabular}
    }\\
     \subfloat[][Synapse]
     {  
        \scriptsize
    	\begin{tabular}{ccc} \toprule
        &nnFormer& nnUNet\\
        \hline
        Spacing &$[0.76,\ 0.76,\ 3]$&$[0.76,\ 0.76,\ 3]$\\
        Median shape &$512\times512\times148$&$512\times512\times148$\\
        Crop size &$128\times128\times64$&$192\times192\times48$\\
        Batch size &2&2\\
        DS Str. &\makecell[c]{$[2,2,2],[2,2,1],[2,2,2]$,\\$[2,2,2],[2,2,2]$} & \makecell[c]{$[2,2,1],[2,2,2],[2,2,2]$,\\$[2,2,2],[2,2,1]$}\\
        \bottomrule
    	\end{tabular}
     }
     \\
     \subfloat[][ACDC]
     {
        \scriptsize
    	\begin{tabular}{ccc} \toprule
        &nnFormer& nnUNet\\
        \hline
        Spacing &$[1.52,\ 1.52,\ 6.35]$&$[1.52,\ 1.52,\ 6.35]$\\
        Median shape &$246\times213\times13$&$246\times213\times13$\\
        Crop size &$160\times160\times14$&$256\times224\times14$\\
        Batch size &4&4\\
        DS Str. &\makecell[c]{$[2,2,1],[2,2,1],[2,2,1]$,\\$[2,2,2],[2,2,2]$} &\makecell[c]{$[2,2,1],[2,2,1],[2,2,2]$,\\$[2,2,1],[2,2,1]$}\\
        \bottomrule
    	\end{tabular}
     }
    %  \subfloat[][Cerebral hemorrhage]
    %  {
    %     \scriptsize
    % 	\begin{tabular}{ccc} \toprule
    %     &nnFormer& nnUNet\\
    %     \hline
    %     Spacing &$[1.0,\ 1.0,\ 1.0]$&$[1.0,\ 1.0,\ 1.0]$\\
    %     Median shape &$397\times419\times28$&$397\times419\times28$\\
    %     Crop size &$336\times352\times20$&$336\times352\times20$\\
    %     Batch size &2&2\\
    %     DS Str. &\makecell[c]{$[2,2,2],[2,2,2],[2,2,1]$} &\makecell[c]{$[2,2,2],[2,2,2],[2,2,1]$}\\
    %     \bottomrule
    % 	\end{tabular}
    % }
    \caption{Network configurations of our nnFormer and nnUNet on three public datasets. We only report the down-sampling stride (abbreviated as DS Str.) as the corresponding up-sampling stride can be easily inferred according to symmetrical down-sampling operations. Note that the network configuration of nnUNet is automatically determined based on pre-defined hand-crafted rules (for self-adaptation).}
    \label{net_conf}
\end{table}

\noindent \textbf{The down-sampling layer.} We found that by replacing the patch merging operation in \cite{liu2021swin} with straightforward strided convolution, nnFormer can provide more improvements on volumetric image segmentation. The intuition behind is that convolutional down-sampling produces hierarchical representations that help model object concepts at multiple scales. As displayed in Figure \ref{overview}c, in most cases, the down-sampling layer involves a strided convolution operation where the stride is set to 2 in all dimensions. However, in practice, the stride with respect to specific dimension can be set to 1 as the number of slices is limited in this dimension and over-down-sampling (i.e., using a large down-sampling stride) can be harmful.

\subsection{Bottleneck}
The original vision transformer (i.e., ViT) \cite{dosovitskiy2020image} employs the naive 2D multi-head self-attention mechanism. In this paper, we extend it to a 3D version (as shown in Figure~\ref{fig_gv-msa}), whose computational complexity can be formulated as follows:
\begin{align}
    \begin{split}
        \Omega(\text{GV-MSA}) = 4hwd C^2 + 2(hwd)^2 C.
    \end{split}
\end{align}
Compared to (\ref{formula_lv-msa}), it is obvious that GV-MSA requires much more computational resources when $\{h, w, d\}$ are relatively larger (e.g., an order of magnitude larger) than $\{S_H, S_W, S_D\}$. In fact, this is exactly the reason why we use local transformer blocks in the encoder, which are designed to handle large-sized inputs efficiently with the local self-attention mechanism. 

However, in the bottleneck, $\{h, w, d\}$ already become much smaller after several down-sampling layers, making the product of them, i.e. $hwd$, , have a similar size to that of $S_H S_W S_D$. This creates the condition for applying GV-MSA, which is able to provide larger receptive field compared to LV-MSA and large receptive field has been proven to be beneficial in different applications~\cite{luo2016understanding,liu2018receptive,chen2017deeplab,szegedy2015going}. In practice, we use three global transformer blocks (i.e., six GV-MSA layers) in the bottleneck to provide sufficient receptive field to the decoder.

\subsection{Decoder}
The architecture of two transformer blocks in the decoder is highly symmetrical to those in the encoder. In contrast to the down-sampling blocks, we employ strided deconvolution to up-sample low-resolution feature maps to high-resolution ones, which in turn are merged with representations from the encoder via skip attention to capture both semantic and fine-grained information. Similar to up-sampling blocks, the last patch expanding block also takes the deconvolutional operation to produce final mask predictions.\\

\noindent \textbf{Skip Attention.} Atypical skip connections in convnets~\cite{ronneberger2015u,he2016deep} adapt either concatenation or summation to incorporate more information. Inspired by the machine translation task in~\cite{vaswani2017attention}, we propose to replace the concatenation/summation with an attention mechanism, which is named as Skip Attention in this paper. To be specific, the output of the $l$-th transformer block of the encoder, i.e., $\mathcal{X}_{\{\text{LV}, \text{GV}\}}^l$, is transformed and split into a key matrix $K^{l^*}$ and a value matrix $V^{l^*}$ after the linear projection (i.e, a one-layer neural network):
\begin{align}
    \begin{split}
        K^{l^*}, V^{l^*} = \text{LP}(\mathcal{X}_{\{\text{LV}, \text{GV}\}}^l),
    \end{split}
\end{align}
where LP stands for the linear projection. Accordingly, $\mathcal{X}_{\text{UP}}^{l^*}$, the output feature maps after the $l^*$-th up-sampling layer of the decoder, is treated as the query $Q^{l^*}$. Then, we can conduct LV/GV-MSA on $Q^{l^*}$, $K^{l^*}$ and $V^{l^*}$ in the decoder like what we have done in (\ref{qkv_op}), i.e.,
\begin{align}
    \small
    \begin{split}
        \text{Attention}(Q^{l^*}, K^{l^*}, V^{l^*})=\text{softmax} \left(\frac{Q^{l^*} (K^{l^*})^{T}}{\sqrt{d^{l^*}_k}}+B^{l^*}\right) V^{l^*},
    \end{split}
    \label{qkv_op2}
\end{align}
where $l^*$ denotes the layer index. $d^{l^*}_k$ and $B^{l^*}$ have the same meaning as those in (\ref{qkv_op}), whose sizes can be easily inferred, accordingly.

\section{Experiments}
\label{sec:experiment}
\begin{table*}[!htp]
\begin{center}
    \resizebox{1.0\textwidth}{!}{
    \begin{tabular}{l|cc|cc|cc|cc}
    \toprule
       \multirow{2}{*}{Methods}  & \multicolumn{2}{c|}{Average} & \multicolumn{2}{c|}{WT} & \multicolumn{2}{c|}{ET} & \multicolumn{2}{c}{TC}\\  \cline{2-9}
       &  HD95 $\downarrow$ & DSC $\uparrow$ & HD95 $\downarrow$ & DSC $\uparrow$ & HD95 $\downarrow$ & DSC $\uparrow$ & HD95 $\downarrow$ & DSC $\uparrow$ \\
       \hline
       \hline
       SETR NUP~\cite{zheng2021rethinking} & 13.78 & 63.7 & 14.419 & 69.7 & 11.72 & 54.4 & 15.19 & 66.9 \\
       SETR PUP~\cite{zheng2021rethinking} & 14.01 & 63.8 & 15.245 & 69.6 & 11.76 & 54.9 & 15.023 & 67.0 \\
       SETR MLA~\cite{zheng2021rethinking} & 13.49 & 63.9 & 15.503 & 69.8 & 10.24 & 55.4 & 14.72 & 66.5 \\
       TransUNet~\cite{chen2021transunet} & 12.98 & 64.4 & 14.03 & 70.6 & 10.42 & 54.2 & 14.5 & 68.4 \\
       TransBTS~\cite{wang2021transbts} & 9.65 & 69.6 & 10.03 & 77.9 & 9.97 & 57.4 & 8.95 & 73.5 \\
       CoTr w/o CNN encoder~\cite{xie2021cotr} & 11.22 & 64.4 & 11.49 & 71.2 & 9.59 & 52.3 & 12.58 & 69.8 \\
       CoTr~\cite{xie2021cotr} & 9.70 & 68.3 & 9.20 & 74.6 & 9.45 & 55.7 & 10.45 & 74.8 \\
       UNETR~\cite{Hatamizadeh_2022_WACV} & 8.82 & 71.1 & 8.27 & 78.9 & 9.35 & 58.5 & 8.85 & 76.1 \\
       \hline
       Our nnFormer & \textbf{4.05} & \textbf{86.4} & \textbf{3.80} & \textbf{91.3} & \textbf{3.87} & \textbf{81.8} & \textbf{4.49} & \textbf{86.0}\\
       \hline
       P-values & \multicolumn{8}{c}{$<$ 1e-2 (HD95), $<$ 1e-2 (DSC)} \\
    \bottomrule
    \end{tabular}
    }
    \caption{Comparison with transformer-based models on brain tumor segmentation. The evaluation metrics are HD95 (mm) and DSC in (\%). Best results are bolded while second best are underlined. Experimental results of baselines are from \cite{Hatamizadeh_2022_WACV}. We calculate the p-values between the average performance of our nnFormer and the best performing baseline in both metrics.}
    \label{tb_bts}
\end{center}
\end{table*}
\begin{table*}[htp]
    \begin{center}
	\resizebox{\textwidth}{!}{
	\begin{tabular}{l|cc|c|c|c|c|c|c|c|c} 
	\toprule
	    \multirow{2}{*}{Methods} & \multicolumn{2}{c|}{Average} & \multirow{2}{*}{Aotra} & \multirow{2}{*}{Gallbladder}  & \multirow{2}{*}{Kidney (Left)}  & \multirow{2}{*}{Kidney (Right)}  & \multirow{2}{*}{Liver}  & \multirow{2}{*}{Pancreas} & \multirow{2}{*}{Spleen}  & \multirow{2}{*}{Stomach} \\ \cline{2-3}
	    & HD95 $\downarrow$ & DSC $\uparrow$ &&&&&&&&\\
		\hline
		\hline
% 		  VNet~\cite{milletari2016v}  & - & 68.81  & 75.34  & 51.87  & 77.10  & 80.75  & 87.84  & 40.04  & 80.56  & 56.98 \\
%         DARR~\cite{fu2020domain}  & - & 69.77 & 74.74  & 53.77  & 72.31 & 73.24  & 94.08  & 54.18  & 89.90  & 45.96 \\
%         R50 U-Net~\cite{ronneberger2015u}  & 36.87 & 74.68  & 87.74  & 63.66  & 80.60 & 78.19  & 93.74  & 56.90  & 85.87  & 74.16 \\
%         U-Net~\cite{ronneberger2015u}  & 39.70 & 76.85  & 89.07  & 69.72 & 77.77 & 68.60  & 93.43  & 53.98  & 86.67  & 75.58 \\
%         R50 Att-UNet~\cite{SCHLEMPER2019197}  & 36.97 & 75.57  & 55.92  & 63.91  & 79.20 & 72.71  & 93.56  & 49.37  & 87.19  & 74.95 \\
%         Att-UNet~\cite{oktay2018attention}  & 36.02 & 77.77  & 89.55  & 68.88  & 77.98  & 71.11  & 93.57  & 58.04  & 87.30  & 75.75 \\
		ViT \cite{dosovitskiy2020image} + CUP \cite{chen2021transunet} & 36.11 & 67.86 & 70.19 & 45.10 & 74.70 & 67.40 & 91.32 & 42.00 & 81.75 & 70.44\\
		R50-ViT \cite{dosovitskiy2020image} + CUP \cite{chen2021transunet} & 32.87 & 71.29 & 73.73 & 55.13 & 75.80 & 72.20 & 91.51 & 45.99 & 81.99 & 73.95\\
        TransUNet~\cite{chen2021transunet}	& 31.69 & 77.48 &  87.23 & 63.16 & 81.87 & 77.02 & 94.08 & 55.86 & 85.08 & 75.62\\
        TransUNet{$^\bigtriangledown$}~\cite{chen2021transunet}	& - & \underline{84.36} & \underline{90.68} & \textbf{71.99} & \underline{86.04} & \underline{83.71} & \underline{95.54} & \underline{73.96} & 88.80 & \underline{84.20}\\
        SwinUNet~\cite{cao2021swin}	& 21.55 & 79.13 & 85.47 & 66.53 &83.28 &79.61 & 94.29 & 56.58 & \underline{90.66} & 76.60\\
        TransClaw U-Net~\cite{chang2021transclaw} & 26.38 & 78.09 & 85.87 & 61.38 & 84.83 & 79.36 & 94.28 & 57.65 & 87.74 &73.55\\
        LeVit-UNet-384s~\cite{xu2021levit}	&\underline{16.84}&78.53 &87.33&62.23	&84.61	&80.25	&93.11	&59.07	&88.86	&72.76\\
        MISSFormer~\cite{huang2021missformer} & 18.20 & 81.96 & 86.99 & 68.65 & 85.21 & 82.00 & 94.41 & 65.67 & \textbf{91.92} & 80.81\\
        % \hline
        % nnUNet (3D) \cite{isensee2019automated}	&86.99	&\textbf{93.01}	&\textbf{71.77}	&85.57	&\textbf{88.18}	&\textbf{97.23}	&\textbf{83.01}	&91.86	&85.25\\
        UNETR~\cite{Hatamizadeh_2022_WACV} & 22.97 & 79.56 & 89.99 & 60.56 & 85.66 & 84.80 & 94.46 & 59.25 & 87.81 & 73.99 \\
        \hline
        Our nnFormer & \textbf{10.63} & \textbf{86.57} & \textbf{92.04} & \underline{70.17} & \textbf{86.57} & \textbf{86.25} & \textbf{96.84} & \textbf{83.35} & 90.51 & \textbf{86.83}\\
        \hline
        P-values & \multicolumn{10}{c}{$<$ 1e-2 (HD95), $<$ 1e-2 (DSC)} \\
        % Our nnFormer{$^\star$} & 7.70 & 87.51 & 93.11 & 72.08 & 86.20 & 87.76 & 97.20 & 84.21 & 91.94 & 87.60\\
        \bottomrule
	\end{tabular}}
	\caption{Comparison with transformer-based models on multi-organ segmentation (Synapse). The evaluation metrics are HD95 (mm) and DSC in (\%). Best results are bolded while second best are underlined. $\bigtriangledown$ denotes TransUNet uses larger inputs, whose size is 512$\times$512. The p-values are calculated based on the average performance of our nnFormer and the best performing baseline in both metrics.}
	\label{synapse}
	\end{center}
\end{table*}

\begin{table}[h]
    \begin{center} 
    \resizebox{0.5\textwidth}{!}{
	\begin{tabular}{l|c|ccc} 
	\toprule
		Methods & Average & RV & Myo & LV \\
		\hline
		\hline
        % R50-U-Net~\cite{ronneberger2015u}	&87.55	&87.10	&80.63	&94.92\\
        % R50-Attn UNet~\cite{SCHLEMPER2019197}	&86.75	&87.58	&79.20	&93.47\\
        VIT-CUP~\cite{dosovitskiy2020image}	&81.45	&81.46	&70.71	&92.18\\
        R50-VIT-CUP~\cite{dosovitskiy2020image}	&87.57	&86.07	&81.88	&94.75\\
        % CBAM\cite{woo2018cbam}	&87.30	&87.70	&82.10	&92.20\\
        % ResUNet\cite{gao2021utnet}	&86.90	&86.20	&82.50	&92.20\\
        % Dual-Attn\cite{gao2021utnet}	&87.00	&86.40	&82.30	&92.40\\
        % UTNET\cite{gao2021utnet}	&88.30	&88.20	&83.50	&93.10\\
        TransUNet~\cite{chen2021transunet}	&89.71	&88.86	&84.54	&\underline{95.73}\\
        SwinUNet~\cite{cao2021swin}	&90.00	&88.55	&85.62	&\textbf{95.83}\\
        LeViT-UNet-384s~\cite{xu2021levit}	& \underline{90.32} & \underline{89.55} & \underline{87.64} & 93.76\\
        UNETR~\cite{Hatamizadeh_2022_WACV} & 88.61 & 85.29 & 86.52 & 94.02\\
        \hline
        % nnUNet (3D) \cite{isensee2019automated}	&91.59	&90.25	&\underline{89.10} & 95.41\\
        nnFormer &\textbf{92.06}	&\textbf{90.94}	&\textbf{89.58}	&95.65\\
        \hline
        P-value & \multicolumn{4}{c}{$<$ 1e-2 (DSC)} \\
	\bottomrule
	\end{tabular}
	}
	\caption{Comparison with transformer-based models on automatic cardiac diagnosis (ACDC). The evaluation metric is DSC (\%). Best results are bolded while second best are underlined. The default evaluation metric is DSC, based on which we calculate the p-value.}
	\label{ACDC}
	\end{center}
\end{table}

For thoroughly comparing nnFormer to previous convnet- and transformer-based architecture, we conduct experiments on three datasets/tasks: the brain tumor segmentation task in Medical Segmentation Decathlon (MSD)~\cite{antonelli2021medical}, Synapse multi-organ segmentation~\cite{landman2015miccai} and Automatic Cardiac Diagnosis Challenge (ACDC)~\cite{bernard2018deep}. For each experiment, we repeat it for ten times and report their average results. We also calculate p-values to demonstrate the significance of nnFormer.\\

\noindent \textbf{Brain tumor segmentation using MRI scans.} This task consists of 484 MRI images, each of which includes four channels, i.e., FLAIR, T1w, T1gd and T2w. The
data was acquired from 19 different institutions and contained a subset of the data used in the
2016 and 2017 Brain Tumor Segmentation (BraTS) challenges~\cite{menze2014multimodal}. The corresponding target ROIs were the three tumor sub-regions, namely edema (ED), enhancing tumor (ET), and non-enhancing tumor (NET). To be consistent with those results reported in UNETR \cite{Hatamizadeh_2022_WACV}, we display the experimental results of the whole tumor
(WT), enhancing tumor (ET) and tumor core (TC) when comparing our nnFormer with transformer-based models. For the split of data, we follow the instruction of UNETR, where ratios of training/validation/test sets are 80\%, 15\% and 5\%, respectively. As above, we use both HD95 and Dice score as evaluation metrics.\\

\noindent \textbf{Synapse for multi-organ CT segmentation.} This dataset includes 30 cases of abdominal CT scans. Following the split used in \cite{chen2021transunet}, 18 cases are extracted to build the training set while the rest 12 cases are used for testing. We report the model performance evaluated with the 95\% Hausdorff Distance (HD95) and Dice score (DSC) on 8 abdominal organs, which are aorta, gallbladder, spleen, left kidney, right kidney, liver, pancreas and stomach\footnote{Here, we follow the evaluation setting of TransUNet.}.\\

\noindent \textbf{ACDC for automated cardiac diagnosis.} ACDC involves 100 patients, with the cavity of the right ventricle, the myocardium of the left ventricle and the cavity of the left ventricle to be segmented. Each case's labels involve left ventricle (LV), right ventricle (RV) and myocardium (MYO). The dataset is split into 70 training samples, 10 validation samples and 20 test samples. The evaluation metrics include both HD95 and Dice score\footnote{Similar to Synapse, we also follow the evaluation setting of TransUNet.}.\\

% \noindent \textbf{Internal cerebral hemorrhage segmentation dataset.} Our cerebral hemorrhage dataset contains 5,132 CT scans for training and 508 CT scans for testing. 20\% of the training data are randomly sampled as the validation set. There are four sub-types of cerebral hemorrhage, which are intraparenchymal hemorrhage, intraventricular hemorrhage, subarachnoid hemorrhage and epidural hemorrhage. In practice, we report DSC score of each type, respectively. \\

\subsection{Implementation details}
We run all experiments based on Python 3.6, PyTorch 1.8.1 and Ubuntu 18.04. All training procedures have been performed on a single NVIDIA 2080 GPU with 11GB memory. The initial learning rate is set to 0.01 and we employ a ``poly'' decay strategy as described in Equation \ref{lrs}. The default optimizer is SGD where we set the momentum to 0.99. The weight decay is set to 3e-5. We utilize both cross entropy loss and dice loss by simply summing them up. The number of training epochs (i.e., max\_epoch in Equation \ref{lrs}) is 1000 and one epoch contains 250 iterations. The number of heads of multi-head self-attention used in different encoder stages is [6, 12, 24, 48] on Synapse. In the rest two datasets, the number of heads becomes [3, 6, 12, 24].
%\begin{equation}
%\text{LR}=\text{initial_lr} \times %\left(1-\frac{\text{epoch}}{\text{num_epoch}}\right)
%\end{equation}
\begin{align}
    \begin{split}
        \text{lr}=\text{initial\_lr} \times (1-\frac{\text{epoch\_id}}{\text{max\_epoch}})^{0.9}.
    \end{split}
    \label{lrs}
\end{align}
\\

% \begin{table*}[t]
% \newcommand{\tabincell}[2]{\begin{tabular}{@{}#1@{}}#2\end{tabular}}
%     \begin{center} 
% 	\caption{Network configurations on Synapse}\label{tab1}
% 	\begin{tabular}{ccc} \toprule
%     Synapse &nnFormer& nnUNet\\
%     \hline
%     Target spacing:&$[3, 0.76, 0.76]$&$[3, 0.76, 0.76]$\\
%     Median image shape:&$148\times512\times512$&$148\times512\times512$\\
%     Crop size:&$64\times128\times128$&$48\times192\times192$\\
%     Batch size:&2&2\\
%     Down-sampling:&\tabincell{l}{[[2,4,4] ,[2,2,2],\\$[2,2,2],[2,2,2]$]} &\tabincell{l}{[[1,2,2], [2,2,2],[2,2,2],\\$[2,2,2],[1,2,2]$]}\\
%     \hline
% 	\end{tabular}
% 	\end{center}
% \end{table*}
% \begin{table*}[h]
% \newcommand{\tabincell}[2]{\begin{tabular}{@{}#1@{}}#2\end{tabular}}
%     \begin{center} 
% 	\caption{Network configurations on ACDC}\label{tab2}
% 	\begin{tabular}{ccc} \toprule
%     ACDC &nnFormer& nnUNet\\
%     \hline
%     Target spacing:&$[6.35, 1.52, 1.52]$&$[6.35, 1.52, 1.52]$\\
%     Median image shape:&$13\times246\times213$&$13\times246\times213$\\
%     Crop size:&$14\times160\times160$&$14\times256\times224$\\
%     Batch size:&4&4\\
%     Downsampling:&\tabincell{l}{[[1,4,4] ,[1,2,2],\\$[2,2,2],[2,2,2]$]} &\tabincell{l}{[[1,2,2],[1,2,2],[2,2,2],\\$[1,2,2],[1,2,2]$]}\\
%     \hline
% 	\end{tabular}
% 	\end{center}
% \end{table*}

\noindent \textbf{Pre-processing and augmentation strategies.} All images will be first resampled to the same target spacing. Augmentations such as rotation, scaling, gaussian noise, gaussian blur, brightness and contrast adjust, simulation of low resolution, gamma augmentation and mirroring are applied in the given order during the training process. \\

\noindent \textbf{Deep supervision.} We also add deep supervision during the training stage. Specifically, the output of each stage in the decoder is passed to the final expanding block, where cross entropy loss and dice loss would be applied. In practice, given the prediction of one typical stage, we down-sample the ground truth segmentation mask to match the prediction's resolution. Thus, the final training objective function is the sum of all losses at three resolutions: 
\begin{align}
    \begin{split}
        \mathcal{L}_{all} = \alpha_1 \mathcal{L}_{\{H,\ W,\ D\}} + \alpha_2 \mathcal{L}_{\{\frac{H}{4},\ \frac{W}{4},\ \frac{D}{2}\}} + \alpha_3 \mathcal{L}_{\{\frac{H}{8},\ \frac{W}{8},\ \frac{D}{4}\}}.
    \end{split}
\end{align}
Here, $\alpha_{\{1,\ 2,\ 3\}}$ denote the magnitude factors for losses in different resolutions. In practice, $\alpha_{\{1,\ 2,\ 3\}}$ halve with each decrease in
resolution, leading to $\alpha_2 = \frac{\alpha_1}{2}$ and $\alpha_3 = \frac{\alpha_1}{4}$. Finally, all weight factors are normalized to 1. \\

% \noindent \textbf{Pre-trained model weights.} Pre-training can be important to provide generalized and transferable representations for downstream tasks. Given the fact that most operations in nnFormer operate on 1D sequences, we explore the possibility of transfering pre-trained weights on natural images to the medical imaging field. More concretely, we aim to reap the benefit of pre-trained weights of MLP layers and QKV attention on ImageNet pre-training. To this goal, we align channel numbers of transformer blocks to those of pre-trained models so that we load the weights of MLP layers and QKV attention. Besides, considering architecture of the encoder and decoder are highly symmetrical, we propose \emph{symmetrical initialization} to reuse the pre-trained weights of the encoder in the decoder. Specifically, transformer blocks with the same input and output resolution are initialized using the same set of model weights (i.e., symmetrical transformer blocks of the encoder and decoder in Figure \ref{overview}).\\

\noindent \textbf{Network configurations.} In Table \ref{net_conf}, we display network configurations of experiments on all three datasets. Compared to nnUNet, in nnFormer, better segmentation results can be achieved with smaller-sized input patches.

\begin{table*}[t]
\centering
\subfloat[Brain tumor segmentation]{
\resizebox{0.95\textwidth}{!}{
\begin{tabular}{l|cc|cc|cc|cc|cc|cc}
     \toprule
     \multirow{2}{*}{Methods}& \multicolumn{2}{c|}{Average} & \multicolumn{2}{c|}{WT} & \multicolumn{2}{c|}{ET} & \multicolumn{2}{c|}{TC} & \multicolumn{2}{c|}{ED} & \multicolumn{2}{c}{NET} \\ \cline{2-13}
     & HD95 $\downarrow$ & DSC $\uparrow$ & HD95 $\downarrow$ & DSC $\uparrow$ & HD95 $\downarrow$ & DSC $\uparrow$ & HD95 $\downarrow$ & DSC $\uparrow$ & HD95 $\downarrow$ & DSC $\uparrow$ & HD95 $\downarrow$ & DSC $\uparrow$  \\
     \hline
     \hline
     nnUNet~\cite{isensee2019automated} & 4.60 & 81.87 & \textbf{3.64} & \textbf{91.99} & 4.06 & 80.97 & 4.91 & 85.35 & 4.26 & \textbf{84.39} & 6.14 & 66.65 \\
     Our nnFormer & \textbf{4.42} & \textbf{82.02} & 3.80 & 91.26 & \textbf{3.87} & \textbf{81.80} & \textbf{4.49} & \textbf{86.02} & \textbf{4.17} & 83.76 & \textbf{5.76} & \textbf{67.29}\\
     \hline
     P-values & \multicolumn{12}{c}{$<$ 1e-2 (HD95), 8.8e-2 (DSC)} \\
     \hline
     nnAvg & \color{ForestGreen}{\textbf{4.09}} & \color{ForestGreen}{\textbf{82.65}} & \color{ForestGreen}{\textbf{3.43}} & \color{ForestGreen}{\textbf{92.33}} & \color{ForestGreen}{\textbf{3.69}} & \color{ForestGreen}{\textbf{82.26}} & \color{ForestGreen}{\textbf{4.17}} & \color{ForestGreen}{\textbf{86.14}} & \color{ForestGreen}{\textbf{3.92}} & \color{ForestGreen}{\textbf{84.95}} & \color{ForestGreen}{\textbf{5.23}} & \color{ForestGreen}{\textbf{67.55}}\\
     \bottomrule
\end{tabular}
}
}\\
\subfloat[Multi-organ segmentation (Synapse)]{
\resizebox{1.0\textwidth}{!}{
\setlength\tabcolsep{1pt}
\begin{tabular}{l|cc|cc|cc|cc|cc|cc|cc|cc|cc}
     \toprule
	 \multirow{2}{*}{Methods} & \multicolumn{2}{c|}{Average} & \multicolumn{2}{c|}{Aotra} & \multicolumn{2}{c|}{Gallbladder}  & \multicolumn{2}{c|}{Kidney (Left)}  & \multicolumn{2}{c|}{Kidney (Right)}  & \multicolumn{2}{c|}{Liver}  & \multicolumn{2}{c|}{Pancreas} & \multicolumn{2}{c|}{Spleen} & \multicolumn{2}{c}{Stomach} \\ \cline{2-19}
	 & HD95 $\downarrow$ & DSC $\uparrow$ & HD95 $\downarrow$ & DSC $\uparrow$ & HD95 $\downarrow$ & DSC $\uparrow$ & HD95 $\downarrow$ & DSC $\uparrow$ & HD95 $\downarrow$ & DSC $\uparrow$ & HD95 $\downarrow$ & DSC $\uparrow$ & HD95 $\downarrow$ & DSC $\uparrow$ & HD95 $\downarrow$ & DSC $\uparrow$ & HD95 $\downarrow$ & DSC $\uparrow$ \\
	 \hline
	 \hline
     nnUNet~\cite{isensee2019automated} & 10.78 & \textbf{86.99} & \textbf{5.91} & \textbf{93.01} & 15.19 & \textbf{71.77} & 18.60 & 85.57 & \textbf{6.44} & \textbf{88.18} & \textbf{1.62} & \textbf{97.23} & 4.52 & 83.01 & 24.34 & \textbf{91.86} & 9.58 & 85.26\\
     Our nnFormer & \textbf{10.63} & 86.57 & 11.38 & 92.04 & \textbf{11.55} & 70.17 & \textbf{18.09} & \textbf{86.57} & 12.76 & 86.25 & 2.00 & 96.84 & \textbf{3.72} & \textbf{83.35} & \textbf{16.92} & 90.51 & \textbf{8.58} & \textbf{86.83} \\
     \hline
     P-values & \multicolumn{18}{c}{2e-2 (HD95), 7.7e-2 (DSC)} \\
     \hline
     nnAvg & \color{ForestGreen}{\textbf{7.70}} & \color{ForestGreen}{\textbf{87.51}} & \color{ForestGreen}{\textbf{5.90}} & \color{ForestGreen}{\textbf{93.11}} & \color{ForestGreen}{\textbf{8.63}} & \color{ForestGreen}{\textbf{72.08}} & {{18.42}} & {{86.20}} & {{8.56}} & {{87.76}} & {1.63} & {{97.20}} & \color{ForestGreen}{\textbf{3.64}} & \color{ForestGreen}{\textbf{84.21}} & \color{ForestGreen}{\textbf{9.42}} & \color{ForestGreen}{\textbf{91.94}} & \color{ForestGreen}{\textbf{5.41}} & \color{ForestGreen}{\textbf{87.60}} \\
     \bottomrule
\end{tabular}
}
}\\
\subfloat[Automated cardiac diagnosis (ACDC)]{
\resizebox{0.7\textwidth}{!}{
\begin{tabular}{c|cc|cc|cc|cc}
     \toprule
     \multirow{2}{*}{Methods} & \multicolumn{2}{c|}{Average} & \multicolumn{2}{c|}{RV} & \multicolumn{2}{c|}{Myo} & \multicolumn{2}{c}{LV}  \\ \cline{2-9}
     & HD95 $\downarrow$ & DSC $\uparrow$ & HD95 $\downarrow$ & DSC $\uparrow$ & HD95 $\downarrow$ & DSC $\uparrow$ & HD95 $\downarrow$ & DSC $\uparrow$ \\
     \hline
     \hline
     nnUNet~\cite{isensee2019automated} & 1.15 & 91.61 & 1.31 & 90.24 & 1.06 & 89.24 & 1.09 & 95.36 \\
     Our nnFormer & \textbf{1.12} & \textbf{92.06} & \textbf{1.23} & \textbf{90.94} & \textbf{1.04} & \textbf{89.58} & 1.09 & \textbf{95.65} \\
     \hline
     P-values & \multicolumn{8}{c}{2e-2 (HD95), $<$ 1e-2 (DSC)} \\
     \hline
     nnAvg & \color{ForestGreen}{\textbf{1.10}} & \color{ForestGreen}{\textbf{92.15}} & \color{ForestGreen}{\textbf{1.19}} & \color{ForestGreen}{\textbf{91.03}} & {{1.04}} & \color{ForestGreen}{\textbf{89.75}} & \color{ForestGreen}{\textbf{1.06}} & \color{ForestGreen}{\textbf{95.68}} \\
     \bottomrule
\end{tabular}
}
}
% \subfloat[Our internal dataset for cerebral hemorrhage segmentation]{
% \resizebox{0.55\textwidth}{!}{
% \begin{tabular}{c|c|c|c|c|c}
%      \toprule
%      Methods& Average & Intraparenchymal & Intraventricular & Subarachnoid & Epidural\\
%      \hline
%      \hline
%      nnUNet~\cite{isensee2019automated} & 57.8 & \textbf{66.0} & 60.0 & \textbf{82.0} & 23.0 \\
%      Our nnFormer & \textbf{62.3} & 65.0 & \textbf{65.0} & 77.0 & \textbf{42.0} \\
%      \hline
%      P-value & \multicolumn{4}{c}{$<$ 1e-2 (DSC)} \\
%      \bottomrule
% \end{tabular}
% }
% }
\caption{Comparison with nnUNet on three public datasets. \textbf{nnAvg} means that we simply average the predictions of nnUNet and nnFormer. Color {\color{ForestGreen}{green}} denotes the target result of nnAvg is the best among all three approaches. Besides, we also highlight the best results between nnUNet and nnFormer in bold font. We calculate p-values between the average performance of nnUNet and our nnFormer in both metrics on three public datasets.}
\label{tb_nnunet}
\end{table*}
\begin{table*}[t]
    \centering
    \resizebox{0.98\textwidth}{!}{
    \begin{tabular}{clcccc}
         \toprule
         \# & Models & Average & RV & Myo & LV \\
         \hline
         \hline
         {\color{gray}{0}} & 1$\times$LV-MSA + PM~\cite{liu2021swin} + PE~\cite{liu2021swin} & 90.55 & 88.59 & 88.47 & 94.60\\
         \hline
         {\color{gray}{1}} & 1$\times$LV-MSA + PM~\cite{liu2021swin} + Conv. Embed. & 90.97 & 88.94 & 88.84 & 95.13 \\
         {\color{gray}{2}} & 1$\times$LV-MSA + Conv. Down. + Conv. Embed. & 91.26 & 89.70 & 89.04 & 95.04 \\
         {\color{gray}{3}} & 1$\times$LV-MSA + 1$\times$GV-MSA + Conv. Down. + Conv. Embed. & 91.46 & 89.82 & 89.17 & 95.39 \\
         {\color{gray}{4}} & 1$\times$LV-MSA + 1$\times$GV-MSA + Conv. Down. + Conv. Embed. + Skip Att. & 91.85 & 90.41 & 89.50 & 95.63 \\
         \hline
         {\color{gray}{5}} & 1$\times$LV-MSA + 1$\times$SLV-MSA + 2$\times$GV-MSA + Conv. Down. + Conv. Embed. + Skip Att. & 92.06 & 90.94 & 89.58 & 95.65 \\
         \bottomrule
    \end{tabular}
    }
    \caption{Investigation of the impact of different modules used in nnFormer. \textbf{PM} and \textbf{PE} denote the patch merging and patch embedding strategies used in swin transformer~\cite{liu2021swin}. \textbf{Conv. Embed.} and \textbf{Conv. Down.} represent our convolutional embedding and down-sampling layers, respectively. \textbf{Skip Att.} refers to the proposed skip attention mechanism. 1$\times$LV-MSA in lines 0-2 means that each transformer block contains one transformer layer and each layer consists of one LV-MSA. 1$\times$GV-MSA in lines 3-4 denotes that we replace LV-MSA in the bottleneck with GV-MSA. 1$\times$SLV-MSA and 2$\times$GV-MSA in line 5 mean that we increase the number of transformer layers in each transformer block from one to two. To be specific, in the encoder/decoder, each transformer block contains 1$\times$LV-MSA and 1$\times$SLV-MSA while in the bottleneck, there are 2$\times$GV-MSA in each block.}
    \label{tb_abs}
\end{table*}
\begin{figure}[t]
    \centering
    \subfloat[Brain tumor segmentation]{\includegraphics[width=0.45\textwidth]{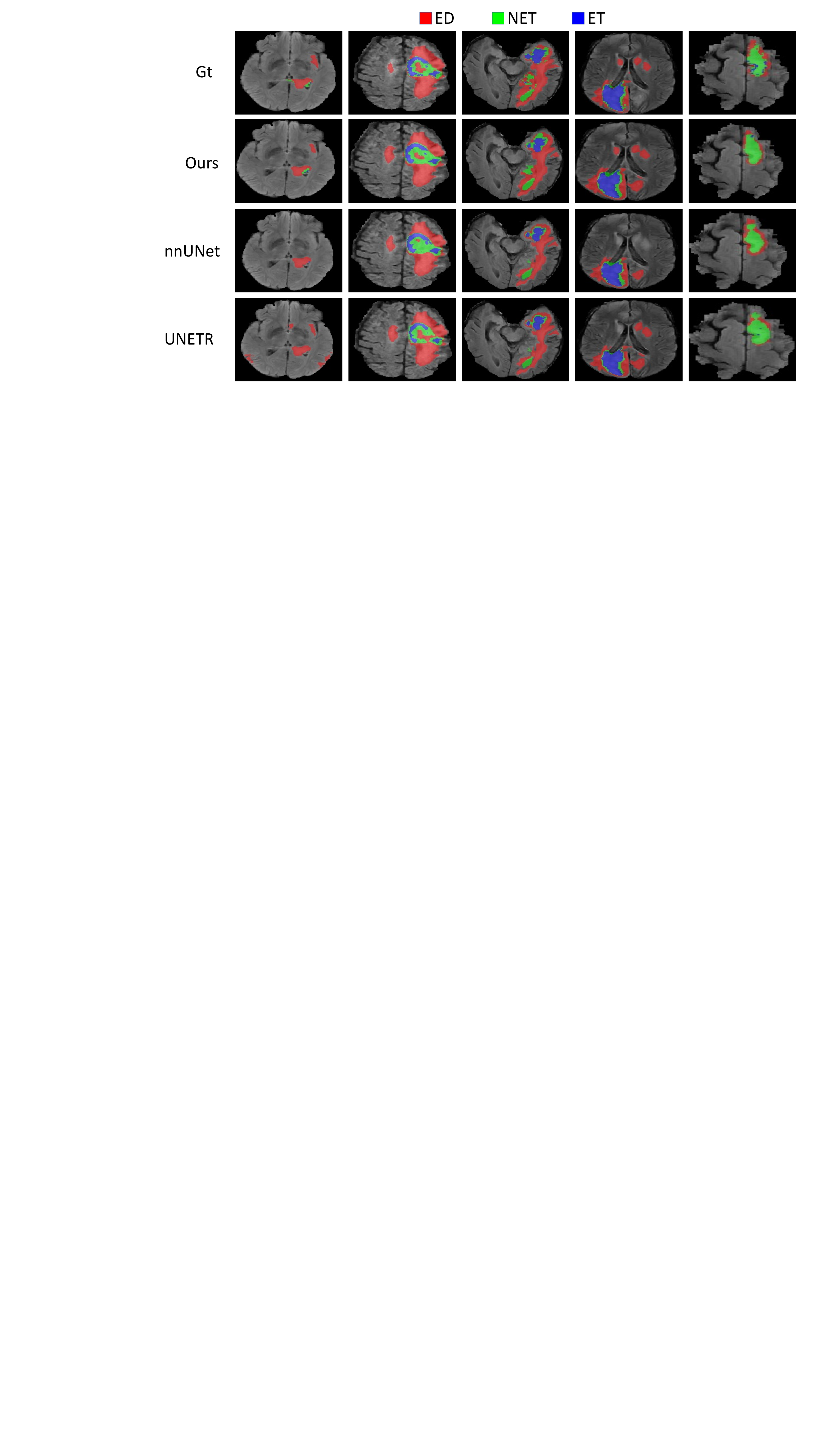}}\\
    \subfloat[Multi-organ segmentation (Synapse)]{\includegraphics[width=0.45\textwidth]{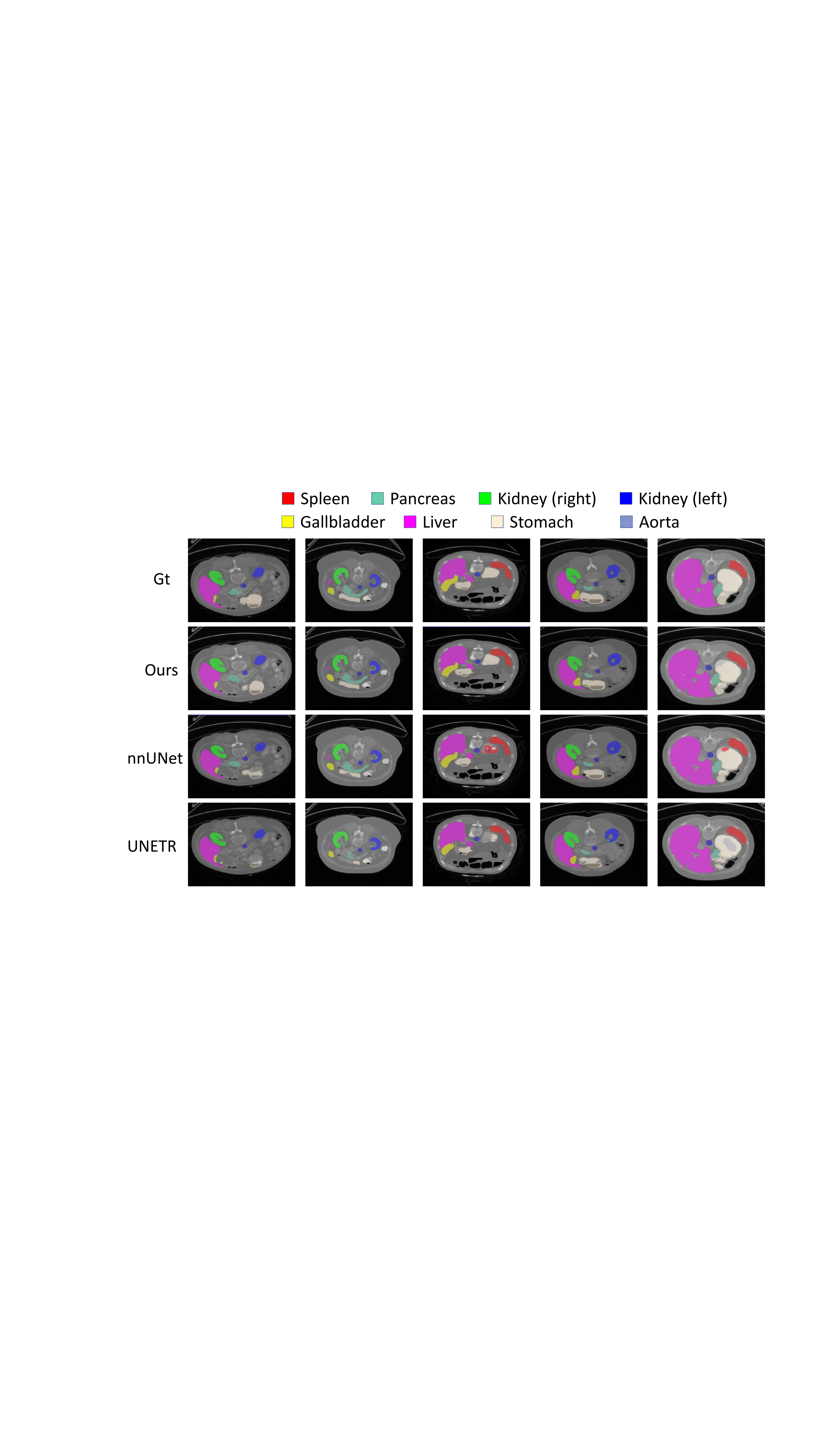}}\\
    \subfloat[Automatic cardiac diagnosis (ACDC)]{\includegraphics[width=0.45\textwidth]{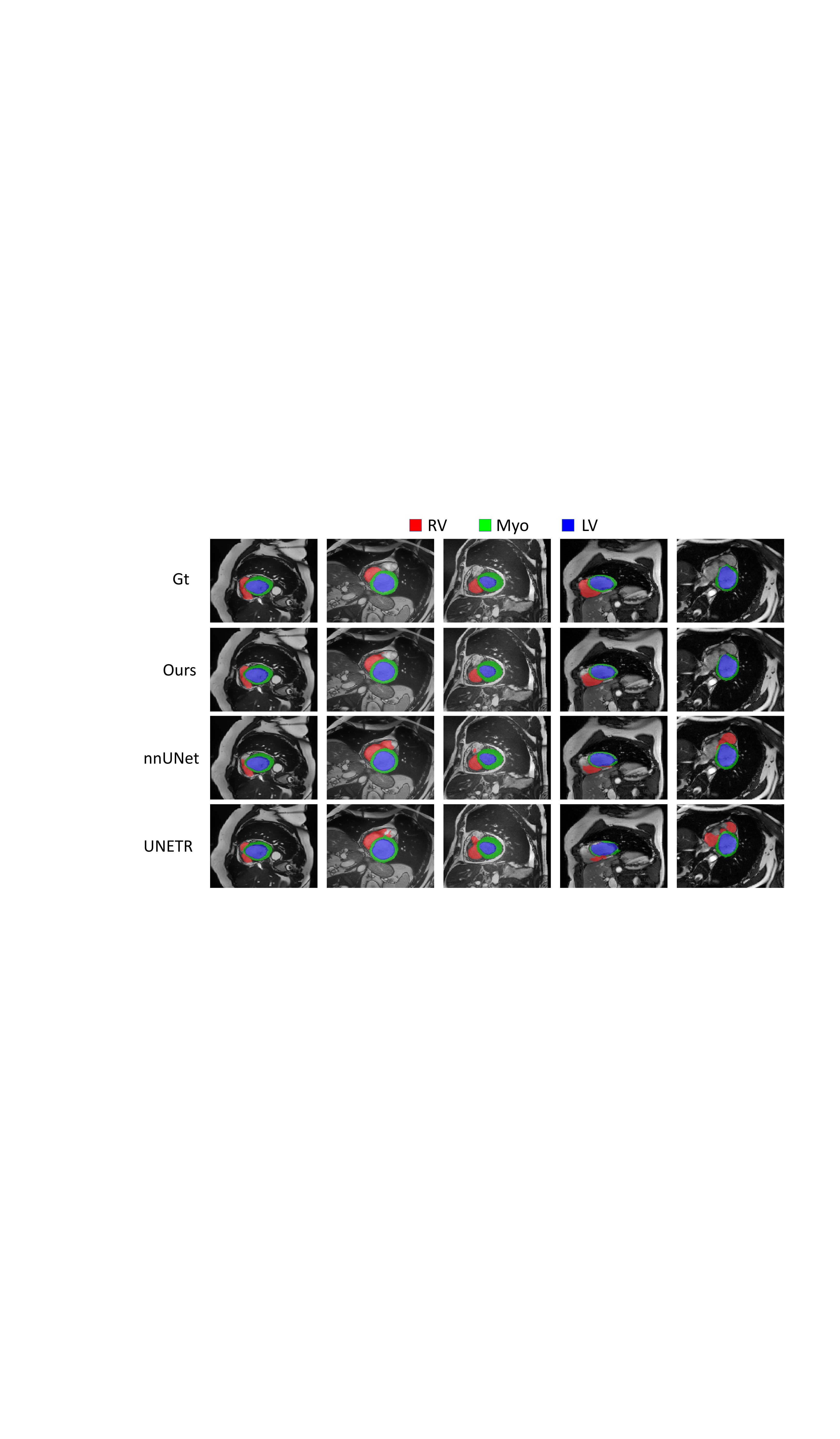}}
    \caption{Visualization of segmentation results on three well-established datasets. We mainly compare nnFormer against nnUNet and UNETR. In addition to segmentation results, we also provide ground truth masks for better comparison.}
    \label{fig_vis}
\end{figure}

\subsection{Comparison with transformer-based methodologies}
\noindent \textbf{Brain tumor segmentation.} Table~\ref{tb_bts} presents experimental results of all models on the task of brain tumor segmentation. Our nnFormer achieves the lowest HD95 and the highest DSC scores in all classes. Moreover, nnFormer is able to surpass the second best method, i.e., UNETR, by large margins in both evaluation metrics. For instance, nnFormer outperforms UNETR by over 4.5 mm in average HD95 and nearly 10 percents in DSC of each class. In comparison to previous transformer-based methods, nnFormer shows more strength in HD95 than in DSC.\\

\noindent \textbf{Multi-organ segmentation (Synapse).} As shown in Table \ref{synapse}, we make experiments on Synapse and to compare our nnFormer against a variety of transformer-based approaches. As we can see, the best performing methods are LeViT-UNet-384s~\cite{xu2021levit} and TransUNet~\cite{chen2021transunet}. LeViT-UNet-384s achieves an average HD95 of 16.84 mm while TransUNet produces an average DSC of 84.36\%. In comparison, our nnFormer is able to outperform LeViT-UNet-384s and TransUNet by over 6 mm and 2 percents in average HD95 and DSC, respectively, which are quite impressive improvements on Synapse. To be specific, nnFormer achieves the highest DSC in six organs, including aotra, kidney (left), kidney (right), liver, pancreas and stomach. Compared to previous transformer-based methods, nnFormer is more advantageous in segmentation pancreas and stomach, both of which are difficult to delineate using past segmentation models.\\

\noindent \textbf{Automated cardiac diagnosis (ACDC).} Table~\ref{ACDC} displays experimental results on ACDC. We can see that the best transformer-based model is LeViT-UNet-384s, whose average DSC is slightly higher than SwinUNet while TransUNet and SwinUNet are more capable of handling the delineation of the left ventricle (LV). In contrast, nnFormer surpasses LeViT-UNet-384s in all classes and by nearly 1.7 percents in average DSC, which again verifies its advantages over past transformer-based approaches.\\

\noindent \textbf{Statistical significance.} In Table~\ref{tb_bts},~\ref{synapse} and~\ref{ACDC}, we employ independent two-sample t-test to calculate p-values between the average performance of our nnFormer and the best performing baseline in both HD95 and DSC. The null hypothesis is that our nnFormer has no advantage over the best performing baseline. As we can see, on all three public datasets, nnFormer produces p-values smaller than 1e-2 under both HD95 and DSC, which indicate strong evidence against the null hypothesis. Thus, nnFormer shows significant improvements over previous transformer-based methods on three different tasks.

\subsection{Comparison with nnUNet and Discussion}
In this section, we compare nnFormer with nnUNet, which has been recognized as one of the most powerful 3D medical image segmentation models~\cite{isensee2019automated}. \\

\noindent \textbf{Results.} In Table~\ref{tb_nnunet}, we display the class-specific results in both HD95 and DSC metrics to make a thorough comparison. To be specific, from the perspective of the class-specific HD95 results, nnFormer outperforms nnUNet in 11 out of 16 categories. In the class-specific DSC, nnFormer outperforms nnUNet in 9 out of 16 categories. Thus, it seems that nnFormer is more advantageous under HD95, which means nnFormer may better delineate the object boundary. From the view of the average performance, we can see that nnFormer often achieves better average performance. For example, nnFormer outperforms nnUNet on all three public datasets with lower HD95 results, while performing better than nnUNet on two out of three datasets with higher DSC results.\\

\noindent \textbf{Statistical significance.} To further verify the significance of nnFormer over nnUNet, we also calculate the p-values between the average performance of nnFormer and nnUNet. Similar to what we have done in Table~\ref{tb_bts}, we provide two p-values based on HD95 and DSC on three public datasets, respectively. The most obvious observation is that nnFormer achieves p-values smaller than 0.05 in HD95 on three public datasets. \emph{These results suggest that nnFormer is the first choice when HD95 is treated as the primary evaluation metric.} Besides, the p-values based on DSC on tumor and multi-organ segmentation ($>$ 0.05) imply that nnFormer is a model comparable to nnUNet, while the results on ACDC demonstrate the significance of nnFormer. In conclusion, \emph{nnFormer has slight advantages over nnUNet under DSC.} \\

\noindent \textbf{Model ensembling.} Besides single model performance, we also investigate the diversity between nnFormer and nnUNet, which is a crucial factor in model ensembling. Somewhat surprisingly, we found that by simply averaging the predictions of nnFormer and nnUNet (i.e., nnAvg in Table~\ref{tb_nnunet}), it can already boost the overall performance by large margins. For instance, nnAvg achieves the best results in all classes under HD95 and DSC on tumor segmentation. Moreover, nnAvg brings nearly 30\% improvements on Synapse when the evaluation metric is HD95. These results indicate that nnFormer and nnUNet are highly complementary to each other.

\subsection{Ablation study}
Table~\ref{tb_abs} displays our ablation study results towards different modules in nnFormer. For simplicity, we made experiments on ACDC and used DSC as the default evaluation metric.

The most basic baseline in Table~\ref{tb_abs} (line 0) consists of LV-MSA (but without SLV-MSA), the patch merging and embedding layers used in \cite{liu2021swin}. We can see that such combination can already achieve a higher average DSC than LeViT-UNet-38~\cite{xu2021levit}, which is the best performing baseline in Table~\ref{ACDC}. We firstly replaced the patch embedding layer, which is implemented with large kernel size and convolutional stride, with our proposed volume embedding layer, i.e., successive convolutional layers with small kernel size and convolutional stride. We found that the introduced convolutional embedding layer improves the average DSC by approximate 0.4 percents. 

Next, we removed the patch merging layer and added our convolutional down-sampling layer. We found such simple replacement can further boost the overall performance by 0.3 percents. Then, we replaced LV-MSA in the bottleneck with GV-MSA, where we observed 0.2-percent improvements. This phenomenon indicates that providing sufficient larger receptive field can be beneficial to the segmentation task. Afterwards, we use skip attention to replace traditional concatenation/summation operations. Somewhat surprisingly, we found that the skip attention is able to boost the overall performance by 0.4 percents, which demonstrates that the skip attention may serve as an alternative choice other than traditional skip connections. Last but not the least, we investigate adding more transformer layers to each transformer block by cascading an SLV-MSA layer with every LV-MSA layer as in Swin Transformer and doubling the number of global self-attention layers. We found that introducing more transformer layers does bring more improvements to the overall performance as it entangles more long-range dependencies into the learned volume representations. 

\subsection{Visualization of segmentation results}
In Figure~\ref{fig_vis}, we visualize some segmentation results of our nnFormer, nnUNet and UNETR on three public datasets. Compared to UNETR, our nnFormer can greatly reduce the number of false positive predictions. One typical example is the fifth example on ACDC. We can see that UNETR produces a large number of wrong right ventricle pixels outside the myocardium. In contrast, our nnFormer generates no prediction of right ventricle outside the myocardium, which demonstrates that nnFormer is more discriminative than UNETR on ACDC. 

On the other hand, we observe that nnUNet displays very competitive segmentation results, much better than UNETR in nearly all examples. However, we still find that nnFormer maintains clear advantages over nnUNet, one of which is that nnFormer is better at dealing with the boundary. In fact, this phenomenon has been reflected in Table~\ref{tb_abs}, where nnFormer is significantly better than nnUNet when HD95 is the default evaluation metric. In Figure~\ref{fig_vis}, we can also observe some evidences. For instance, in the second example on Synapse, nnFormer captures the shape of the left kidney and stomach better than nnUNet. Also, in the third example on brain tumor segmentation, nnUNet misses a major part of the non-enhancing tumor enclosed by the edema. These results verify that our nnFormer has the potential to be treated as an alternative to nnUNet.

\section{Conclusion}
\label{sec:conclusion}
In this paper, we present a 3D transformer, nnFormer, for volumetric image segmentation. nnFormer is constructed on top of an interleaved stem of convolution and self-attention. Convolution helps encode precise spatial information and builds hierarchical object concepts. For self-attention, nnFormer employs three types of attention mechanism to entangle long-range dependencies. Specifically, local and global volume-based self-attention focus on constructing feature pyramids and providing large receptive field. Skip attention is responsible for bridging the gap between the encoder and decoder. Experiments show that nnFormer maintains great advantages over previous transformer-based models in both HD95 and DSC. Compared to nnUNet, nnFormer is significantly better in HD95 while producing comparable results in DSC. More importantly, we demonstrate that nnFormer and nnUNet can be beneficial to each other in model ensembling, where the simple averaging operation can already produce great improvements.

% In this paper, we present a new medical image segmentation network named nnFormer. nnFormer is constructed on top of an interleaved stem of convolution and self-attention, where convolution helps encode precise spatial information into high-resolution low-level features and build hierarchical object concepts at multiple scales. On the other hand, self-attention in transformer blocks entangles long-term dependencies with convolutional representations to capture global context. Based on such hybrid architecture, nnFormer achieves tremendous progress over previous transformer-based segmentation methodologies. Even when compared to nnUNet, currently the best performing segmentation network, nnFormer still provides consistent yet observable improvements. In the future, we hope nnFormer could draw more attention from the medical imaging community to make efforts on developing more efficient segmentation models.

{
\bibliographystyle{ieeetr}
\bibliography{egbib}

\begin{thebibliography}{10}

\bibitem{vaswani2017attention}
A.~Vaswani, N.~Shazeer, N.~Parmar, J.~Uszkoreit, L.~Jones, A.~N. Gomez, {\em
  et~al.}, ``Attention is all you need,'' in {\em Advances in Neural
  Information Processing Systems}, pp.~5998--6008, 2017.

\bibitem{dosovitskiy2020image}
A.~Dosovitskiy, L.~Beyer, A.~Kolesnikov, D.~Weissenborn, X.~Zhai,
  T.~Unterthiner, {\em et~al.}, ``An image is worth 16x16 words: Transformers
  for image recognition at scale,'' {\em arXiv preprint arXiv:2010.11929},
  2020.

\bibitem{liu2021swin}
Z.~Liu, Y.~Lin, Y.~Cao, H.~Hu, Y.~Wei, Z.~Zhang, S.~Lin, and B.~Guo, ``Swin
  transformer: Hierarchical vision transformer using shifted windows,'' {\em
  arXiv preprint arXiv:2103.14030}, 2021.

\bibitem{he2021masked}
K.~He, X.~Chen, S.~Xie, Y.~Li, P.~Doll{\'a}r, and R.~Girshick, ``Masked
  autoencoders are scalable vision learners,'' {\em arXiv preprint
  arXiv:2111.06377}, 2021.

\bibitem{carion2020end}
N.~Carion, F.~Massa, G.~Synnaeve, N.~Usunier, A.~Kirillov, and S.~Zagoruyko,
  ``End-to-end object detection with transformers,'' in {\em European
  Conference on Computer Vision}, pp.~213--229, Springer, 2020.

\bibitem{lecun1998gradient}
Y.~LeCun, L.~Bottou, Y.~Bengio, and P.~Haffner, ``Gradient-based learning
  applied to document recognition,'' {\em Proceedings of the IEEE}, vol.~86,
  no.~11, pp.~2278--2324, 1998.

\bibitem{zhou2021convnets}
H.-Y. Zhou, C.~Lu, S.~Yang, and Y.~Yu, ``{ConvNets vs. Transformers}: Whose
  visual representations are more transferable?,'' {\em ICCV Workshop on Deep
  Multi-Task Learning in Computer Vision}, 2021.

\bibitem{qu2022m3net}
T.~Qu, X.~Wang, C.~Fang, L.~Mao, J.~Li, P.~Li, J.~Qu, X.~Li, H.~Xue, Y.~Yu,
  {\em et~al.}, ``M3net: A multi-scale multi-view framework for multi-phase
  pancreas segmentation based on cross-phase non-local attention,'' {\em
  Medical Image Analysis}, vol.~75, p.~102232, 2022.

\bibitem{zhang2021cross}
D.~Zhang, G.~Huang, Q.~Zhang, J.~Han, J.~Han, and Y.~Yu, ``Cross-modality deep
  feature learning for brain tumor segmentation,'' {\em Pattern Recognition},
  vol.~110, p.~107562, 2021.

\bibitem{tuli2021convolutional}
S.~Tuli, I.~Dasgupta, E.~Grant, and T.~L. Griffiths, ``Are convolutional neural
  networks or transformers more like human vision?,'' {\em arXiv preprint
  arXiv:2105.07197}, 2021.

\bibitem{chen2021transunet}
J.~Chen, Y.~Lu, Q.~Yu, X.~Luo, E.~Adeli, Y.~Wang, {\em et~al.}, ``{TransUNet}:
  Transformers make strong encoders for medical image segmentation,'' {\em
  arXiv preprint arXiv:2102.04306}, 2021.

\bibitem{ronneberger2015u}
O.~Ronneberger, P.~Fischer, and T.~Brox, ``{U-Net}: Convolutional networks for
  biomedical image segmentation,'' in {\em International Conference on Medical
  image computing and computer-assisted intervention}, pp.~234--241, Springer,
  2015.

\bibitem{zhang2021transfuse}
Y.~Zhang, H.~Liu, and Q.~Hu, ``{TransFuse}: Fusing transformers and cnns for
  medical image segmentation,'' {\em arXiv preprint arXiv:2102.08005}, 2021.

\bibitem{valanarasu2021medical}
J.~M.~J. Valanarasu, P.~Oza, I.~Hacihaliloglu, and V.~M. Patel, ``{Medical
  Transformer}: Gated axial-attention for medical image segmentation,'' {\em
  arXiv preprint arXiv:2102.10662}, 2021.

\bibitem{chang2021transclaw}
Y.~Chang, H.~Menghan, Z.~Guangtao, and Z.~Xiao-Ping, ``{TransClaw U-Net}: Claw
  u-net with transformers for medical image segmentation,'' {\em arXiv preprint
  arXiv:2107.05188}, 2021.

\bibitem{chen2021transattunet}
B.~Chen, Y.~Liu, Z.~Zhang, G.~Lu, and D.~Zhang, ``{TransAttUnet}: Multi-level
  attention-guided u-net with transformer for medical image segmentation,''
  {\em arXiv preprint arXiv:2107.05274}, 2021.

\bibitem{karimi2021convolution}
D.~Karimi, S.~Vasylechko, and A.~Gholipour, ``Convolution-free medical image
  segmentation using transformers,'' {\em arXiv preprint arXiv:2102.13645},
  2021.

\bibitem{cao2021swin}
H.~Cao, Y.~Wang, J.~Chen, D.~Jiang, X.~Zhang, Q.~Tian, and M.~Wang,
  ``{Swin-Unet}: Unet-like pure transformer for medical image segmentation,''
  {\em arXiv preprint arXiv:2105.05537}, 2021.

\bibitem{lin2021ds}
A.~Lin, B.~Chen, J.~Xu, Z.~Zhang, and G.~Lu, ``{DS-TransUNet}: Dual swin
  transformer u-net for medical image segmentation,'' {\em arXiv preprint
  arXiv:2106.06716}, 2021.

\bibitem{li2021medical}
S.~Li, X.~Sui, X.~Luo, X.~Xu, Y.~Liu, and R.~S.~M. Goh, ``Medical image
  segmentation using squeeze-and-expansion transformers,'' {\em arXiv preprint
  arXiv:2105.09511}, 2021.

\bibitem{yun2021spectr}
B.~Yun, Y.~Wang, J.~Chen, H.~Wang, W.~Shen, and Q.~Li, ``{SpecTr}: Spectral
  transformer for hyperspectral pathology image segmentation,'' {\em arXiv
  preprint arXiv:2103.03604}, 2021.

\bibitem{xu2021levit}
G.~Xu, X.~Wu, X.~Zhang, and X.~He, ``{LeViT-UNet}: Make faster encoders with
  transformer for medical image segmentation,'' {\em arXiv preprint
  arXiv:2107.08623}, 2021.

\bibitem{li2021more}
Y.~Li, W.~Cai, Y.~Gao, and X.~Hu, ``More than encoder: Introducing transformer
  decoder to upsample,'' {\em arXiv preprint arXiv:2106.10637}, 2021.

\bibitem{xie2021cotr}
Y.~Xie, J.~Zhang, C.~Shen, and Y.~Xia, ``{CoTr}: Efficiently bridging cnn and
  transformer for 3d medical image segmentation,'' {\em arXiv preprint
  arXiv:2103.03024}, 2021.

\bibitem{wang2021transbts}
W.~Wang, C.~Chen, M.~Ding, J.~Li, H.~Yu, and S.~Zha, ``{TransBTS}: Multimodal
  brain tumor segmentation using transformer,'' {\em arXiv preprint
  arXiv:2103.04430}, 2021.

\bibitem{hendrycks2016gaussian}
D.~Hendrycks and K.~Gimpel, ``Gaussian error linear units (gelus),'' {\em arXiv
  preprint arXiv:1606.08415}, 2016.

\bibitem{ba2016layer}
J.~L. Ba, J.~R. Kiros, and G.~E. Hinton, ``Layer normalization,'' {\em arXiv
  preprint arXiv:1607.06450}, 2016.

\bibitem{luo2016understanding}
W.~Luo, Y.~Li, R.~Urtasun, and R.~Zemel, ``Understanding the effective
  receptive field in deep convolutional neural networks,'' in {\em Proceedings
  of the 30th International Conference on Neural Information Processing
  Systems}, pp.~4905--4913, 2016.

\bibitem{liu2018receptive}
S.~Liu, D.~Huang, {\em et~al.}, ``Receptive field block net for accurate and
  fast object detection,'' in {\em Proceedings of the European Conference on
  Computer Vision}, pp.~385--400, 2018.

\bibitem{chen2017deeplab}
L.-C. Chen, G.~Papandreou, I.~Kokkinos, K.~Murphy, and A.~L. Yuille, ``Deeplab:
  Semantic image segmentation with deep convolutional nets, atrous convolution,
  and fully connected crfs,'' {\em IEEE Transactions on Pattern Analysis and
  Machine Intelligence}, vol.~40, no.~4, pp.~834--848, 2017.

\bibitem{szegedy2015going}
C.~Szegedy, W.~Liu, Y.~Jia, P.~Sermanet, S.~Reed, D.~Anguelov, D.~Erhan,
  V.~Vanhoucke, and A.~Rabinovich, ``Going deeper with convolutions,'' in {\em
  Proceedings of the IEEE Conference on Computer Vision and Pattern
  Recognition}, pp.~1--9, 2015.

\bibitem{he2016deep}
K.~He, X.~Zhang, S.~Ren, and J.~Sun, ``Deep residual learning for image
  recognition,'' in {\em Proceedings of the IEEE Conference on Computer Vision
  and Pattern Recognition}, pp.~770--778, 2016.

\bibitem{zheng2021rethinking}
S.~Zheng, J.~Lu, H.~Zhao, X.~Zhu, Z.~Luo, Y.~Wang, Y.~Fu, J.~Feng, T.~Xiang,
  P.~H. Torr, {\em et~al.}, ``Rethinking semantic segmentation from a
  sequence-to-sequence perspective with transformers,'' in {\em Proceedings of
  the IEEE Conference on Computer Vision and Pattern Recognition},
  pp.~6881--6890, 2021.

\bibitem{Hatamizadeh_2022_WACV}
A.~Hatamizadeh, Y.~Tang, V.~Nath, D.~Yang, A.~Myronenko, B.~Landman, H.~R.
  Roth, and D.~Xu, ``{UNETR}: Transformers for 3d medical image segmentation,''
  in {\em Proceedings of the IEEE/CVF Winter Conference on Applications of
  Computer Vision}, pp.~574--584, January 2022.

\bibitem{huang2021missformer}
X.~Huang, Z.~Deng, D.~Li, and X.~Yuan, ``{MISSFormer}: An effective medical
  image segmentation transformer,'' {\em arXiv preprint arXiv:2109.07162},
  2021.

\bibitem{antonelli2021medical}
M.~Antonelli, A.~Reinke, S.~Bakas, K.~Farahani, B.~A. Landman, G.~Litjens,
  B.~Menze, O.~Ronneberger, R.~M. Summers, B.~van Ginneken, {\em et~al.}, ``The
  medical segmentation decathlon,'' {\em arXiv preprint arXiv:2106.05735},
  2021.

\bibitem{landman2015miccai}
B.~Landman, Z.~Xu, J.~E. Igelsias, M.~Styner, T.~Langerak, and A.~Klein,
  ``Miccai multi-atlas labeling beyond the cranial vault--workshop and
  challenge,'' in {\em Proc. MICCAI: Multi-Atlas Labeling Beyond Cranial
  Vault-Workshop Challenge}, 2015.

\bibitem{bernard2018deep}
O.~Bernard, A.~Lalande, C.~Zotti, F.~Cervenansky, X.~Yang, P.-A. Heng,
  I.~Cetin, K.~Lekadir, O.~Camara, M.~A.~G. Ballester, {\em et~al.}, ``Deep
  learning techniques for automatic mri cardiac multi-structures segmentation
  and diagnosis: Is the problem solved?,'' {\em IEEE Transactions on Medical
  Imaging}, vol.~37, no.~11, pp.~2514--2525, 2018.

\bibitem{menze2014multimodal}
B.~H. Menze, A.~Jakab, S.~Bauer, J.~Kalpathy-Cramer, K.~Farahani, J.~Kirby,
  Y.~Burren, N.~Porz, J.~Slotboom, R.~Wiest, {\em et~al.}, ``The multimodal
  brain tumor image segmentation benchmark (brats),'' {\em IEEE Transactions on
  Medical Imaging}, vol.~34, no.~10, pp.~1993--2024, 2014.

\bibitem{isensee2019automated}
F.~Isensee, P.~F. Jaeger, S.~A. Kohl, J.~Petersen, and K.~H. Maier-Hein,
  ``nnu-net: a self-configuring method for deep learning-based biomedical image
  segmentation,'' {\em Nature methods}, vol.~18, no.~2, pp.~203--211, 2021.

\end{thebibliography}
}
\end{document}